\documentclass[10pt,journal,compsoc]{IEEEtran}

\usepackage{amsmath}
\usepackage{graphicx}
\usepackage{amssymb}
\usepackage{pdflscape}
\usepackage{booktabs}
\usepackage{bm}
\usepackage{subfigure}
\usepackage{multirow}
\usepackage{graphicx}
\usepackage[numbers,sort]{natbib}
\usepackage[colorlinks=true, allcolors=blue]{hyperref}

\usepackage{soul}
\usepackage[linguistics]{forest}

\newcommand{\format}[2]{\begin{tabular}{@{}c@{}}$#1$\\[-3pt]{\color{gray}$\scriptscriptstyle \pm #2$}\end{tabular}}

\newcommand{\bformat}[2]{\color{black} \begin{tabular}{@{}c@{}}$\mathbf{#1}$\\[-3pt]$\scriptscriptstyle \mathbf{\pm #2}$\end{tabular}}
\newcommand{\orformat}[2]{\color{orange} \begin{tabular}{@{}c@{}}$\mathbf{#1}$\\[-3pt]$\scriptscriptstyle \mathbf{\pm #2}$\end{tabular}}
\newcommand{\blformat}[2]{\color{blue} \begin{tabular}{@{}c@{}}$\mathbf{#1}$\\[-3pt]$\scriptscriptstyle \mathbf{\pm #2}$\end{tabular}}

\newcommand{\blue}[1]{\begin{color}{blue}#1\end{color}}

\begin{document}
\title{A Comprehensive Survey of Dataset Distillation}

\author{Shiye~Lei
        and~Dacheng~Tao,~\IEEEmembership{Fellow,~IEEE}% <-this % stops a space
%\IEEEcompsocitemizethanks{\IEEEcompsocthanksitem This project was supported by Australian Research Council Project FL-170100117. S. Lei and D. Tao are with the Sydney AI Centre and the School of Computer Science in the Faculty of Engineering at The University of Sydney, 6 Cleveland St, Darlington, NSW 2008, Australia (email: slei5230@uni.sydney.edu.au, dacheng.tao@sydney.edu.au; corresponding author: Dacheng Tao).
%\protect\\
%}% <-this % stops an unwanted space
\thanks{%This project was supported by Australian Research Council Project FL-170100117. \hfil\break 
The authors are with the Sydney AI Centre and the School of Computer Science, Faculty of Engineering, The University of Sydney, Darlington, NSW 2008, Australia (Email: \{slei5230, dacheng.tao\}@sydney.edu.au. Corresponding author: Dacheng Tao).}
}


\markboth{ }%Journal of \LaTeX\ Class Files,~Vol.~14, No.~8, August~2015}%
{Shell \MakeLowercase{\textit{et al.}}: Bare Demo of IEEEtran.cls for Computer Society Journals}



\IEEEtitleabstractindextext{%
\begin{abstract}
Deep learning technology has developed unprecedentedly in the last decade and has become the primary choice in many application domains. This progress is mainly attributed to a systematic collaboration in which rapidly growing computing resources encourage advanced algorithms to deal with massive data. However, it has gradually become challenging to handle the unlimited growth of data with limited computing power. To this end, diverse approaches are proposed to improve data processing efficiency. Dataset distillation, a dataset reduction method, addresses this problem by synthesizing a small typical dataset from substantial data and has attracted much attention from the deep learning community. Existing dataset distillation methods can be taxonomized into meta-learning and data matching frameworks according to whether they explicitly mimic the performance of target data. Although dataset distillation has shown surprising performance in compressing datasets, there are still several limitations such as distilling high-resolution data or data with complex label spaces. This paper provides a holistic understanding of dataset distillation from multiple aspects, including distillation frameworks and algorithms, factorized dataset distillation, performance comparison, and applications. Finally, we discuss challenges and promising directions to further promote future studies on dataset distillation.
\end{abstract}

}

\maketitle

\IEEEdisplaynontitleabstractindextext

% \fi
\IEEEpeerreviewmaketitle

\IEEEraisesectionheading{\section{Introduction}\label{sec:introduction}}

\begin{figure}[t]
    \centering
    \includegraphics[width=0.98\columnwidth]{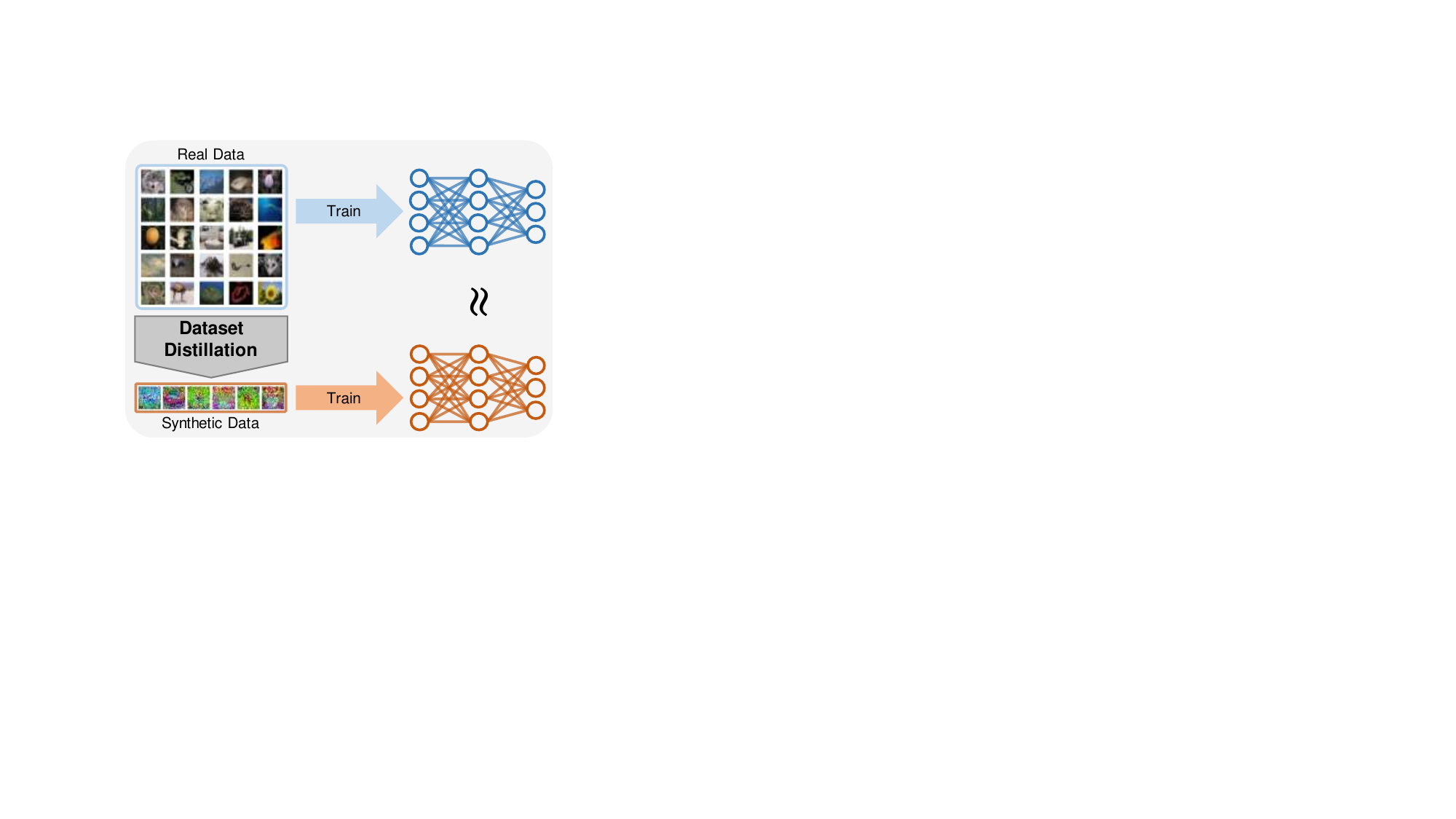}
    \caption{ {An illustration of dataset distillation. The models trained on the large original dataset and small synthetic dataset demonstrate comparable performance on the test set.}}
    \label{fig:dd}
\end{figure}

\IEEEPARstart{D}{uring} the past few decades, deep learning has achieved remarkable success due to powerful computing resources \citep{krizhevsky2017imagenet,he2016deep,vaswani2017attention,chen2020simple,he2020momentum}, which allow deep neural networks to directly handle giant datasets and bypass complicated manual feature extraction \citep{szegedy2016rethinking,szegedy2017inception}. For example, the powerful large language model GPT-3 contains $175$ billion parameters and is trained on $45$ terabytes of text data with thousands of graphics processing units (GPUs) \citep{brown2020language}. However, massive data are generated every day \citep{sagiroglu2013big}, which pose a significant threat to training efficiency and data storage, and deep learning gradually reaches a bottleneck due to the mismatch between the volume of data and computing resources \citep{strubell2019energy}.  {The emergence of {\it dataset distillation} (DD) is precisely to address this problem of large data volume. Dataset distillation, which was first proposed by \citep{wang2018dataset} as a data compression technique that synthesizes a few high-informative data points to summarize numerous real data. Then, downstream models trained on the small synthetic dataset can achieve comparable generalization performance to those trained on the real data. A general illustration for dataset distillation is presented in Figure \ref{fig:dd}.}

\begin{figure*}[t]
    \centering
    \includegraphics[width=1.8\columnwidth]{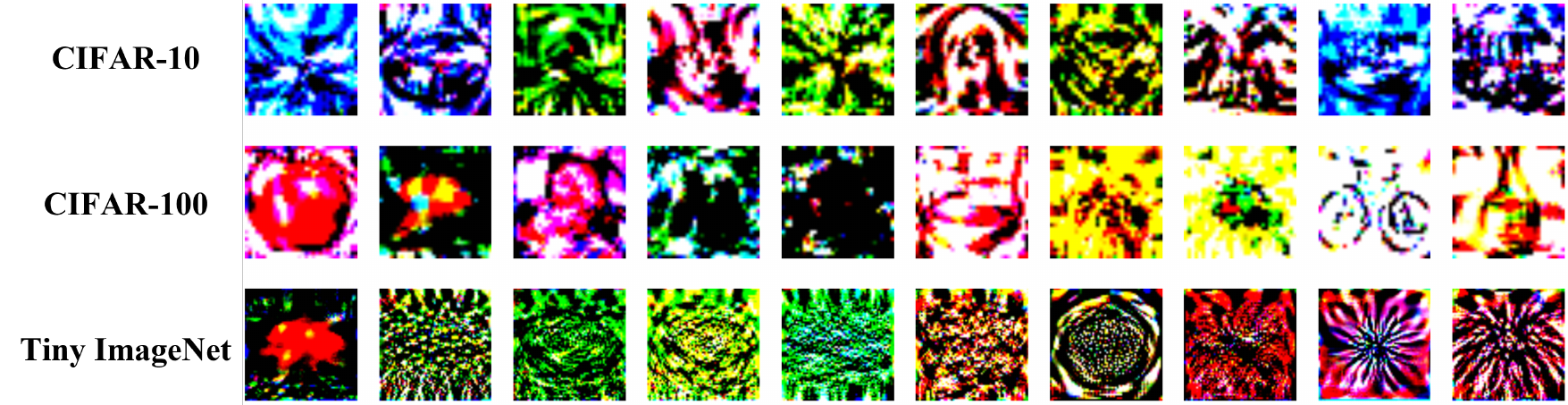}
    \caption{Example synthetic images distilled from CIFAR-10/100 and Tiny ImageNet by matching the training trajectory \citep{cazenavette2022distillation}.}
    \label{fig:dd_sample}
\end{figure*}

Before the appearance of DD, coreset selection \citep{nalepa2019selecting,mirzasoleiman2020coresets,Coleman2020Selection} plays a pivotal role in dataset reduction by selecting a few prototype examples from the original training dataset as the coreset, and then the model is solely trained on the small coreset to save training costs while avoiding large performance drops. However, elements in the coreset are unmodified and constrained by the original data, which considerably restricts the coreset's expressiveness, especially when the coreset budget is limited. Different from coreset selection, dataset distillation removes the restriction of uneditable elements and carefully modifies a small number of examples to preserve more information, as shown in Figure \ref{fig:dd_sample} for synthetic examples. By distilling the knowledge of the large original dataset into a small synthetic set, models trained on the distilled dataset can acquire a better generalization performance compared to the uneditable coreset.

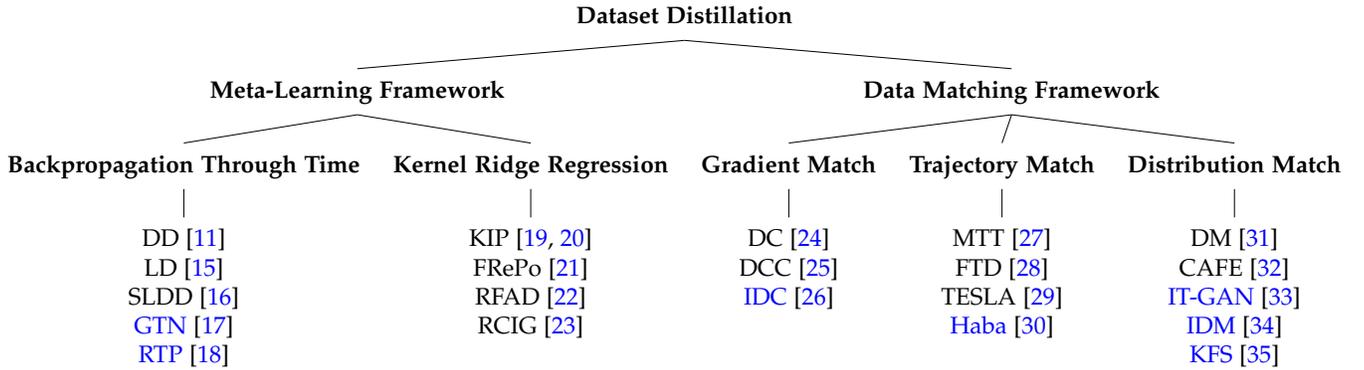
\begin{figure*}[t]
    \centering
    \scalebox{0.95}{
    \begin{forest}
  [\textbf{Dataset Distillation}
    [\textbf{Meta-Learning Framework}
     [\textbf{Backpropagation Through Time}
       [DD \citep{wang2018dataset} \\ LD \citep{bohdal2020flexible} \\ SLDD \citep{sucholutsky2021soft} \\ \blue{GTN} \citep{such2020generative} \\ \blue{RTP} \citep{deng2022remember}]
     ]
     [\textbf{Kernel Ridge Regression}
       [KIP \citep{nguyen2020dataset,nguyen2021dataset} \\ FRePo \citep{zhou2022dataset} \\ RFAD \citep{loo2022efficient} \\ {RCIG} \citep{loo2023dataset}]
     ]
    ]
    [\textbf{Data Matching Framework}
      [\textbf{Gradient Match}
        [DC \citep{zhao2021dataset} \\ DCC \citep{lee2022dataset} \\ \blue{IDC} \citep{kim2022dataset}]
      ]
      [\textbf{Trajectory Match}
        [MTT \citep{cazenavette2022distillation} \\ FTD \citep{du2022minimizing} \\ TESLA \citep{cui2023scaling} \\ \blue{Haba} \citep{liu2022dataset}]
      ]
      [\textbf{Distribution Match}
        [DM \citep{zhao2023distribution} \\ CAFE \citep{wang2022cafe} \\ \blue{IT-GAN} \citep{zhao2022synthesizing} \\ \blue{IDM} \citep{zhao2023improved} \\ \blue{KFS} \citep{lee2022kfs}]
      ]
    ]
  ]
\end{forest}}
    \caption{Tree diagram for different categories of dataset distillation algorithms. The factorized DD methods are marked in blue.}
    \label{fig:tree diagram}
\end{figure*}
Due to the property of extremely high dimensions in the deep learning regime, the data information is hardly disentangled to specific concepts, and thus distilling numerous high-dimensional data into a few points is not a trivial task. Based on the objectives applied to mimic target data, dataset distillation methods can be grouped into meta-learning framework  {(Sec. \ref{sec:meta-learning})} and data matching framework  {(Sec. \ref{sec:data matching})}, and these techniques in each framework can be further classified in a much more detailed manner.  In the meta-learning framework, the distilled data are considered as hyperparameters and optimized in a nested loop fashion according to the distilled-data-trained model's risk {\it w.r.t.} the target data \citep{maclaurin2015gradient,wang2018dataset}. The data matching framework updates distilled data by imitating the influence of target data on model training from parameter or feature space \citep{zhao2021dataset,cazenavette2022distillation,zhao2023distribution}. Figure \ref{fig:tree diagram} presents these different categories of DD algorithms in a tree diagram.

Apart from directly considering the synthetic examples as optimization objectives, in some studies, a proxy model is designed, which consists of latent codes and decoders to generate highly informative examples and to resort to learning the latent codes and decoders  {(Sec. \ref{sec:factorzied dd})}. For example, \citet{such2020generative} employed a network to generate highly informative data from noise and optimized the network with the meta-learning framework. \citet{zhao2022synthesizing} optimized the vectors and put them into the generator of a well-trained generative adversarial network (GAN) to produce the synthetic examples. Moreover, \citet{kim2022dataset,deng2022remember,liu2022dataset,lee2022kfs} learned a couple of latent codes and decoders, and then synthetic data were generated according to the different combinations of latent codes and decodes. With this factorization of synthetic data, the compression ratio of DD can be further decreased, and the performance can also be improved due to the intraclass information extracted by latent codes.

In this paper, we present a comprehensive survey on dataset distillation, and the main objectives of this survey are as follows: (1) present a clear and systematic overview of dataset distillation; (2) review the recent advances in DD algorithms and discuss various applications in dataset distillation; (3) give a comprehensive performance comparison {\it w.r.t.} different dataset distillation algorithms; and (4) provide a detailed discussion on the limitation and promising directions of dataset distillation to benefit future studies.  {We note that there is a concurrent survey on dataset distillation \citep{sachdeva2023data}. The most notable difference between \citep{sachdeva2023data} and our survey is the taxonomy of DD. It plainly classifies DD algorithms into four different categories of meta-model matching, gradient matching, trajectory matching, and distribution matching, while our hierarchical taxonomy first introduces meta-learning and data matching frameworks, which are then followed by a more fine-grained division. Therefore, our taxonomy provides a more systematic framework to track and summarize the development of DD algorithms and better sheds light on future DD studies.}

The rest of this paper is organized as follows. We first provide some background {\it w.r.t.} DD in Section \ref{sec:background}. DD methods under the meta-learning framework are described and comprehensively analyzed in Section \ref{sec:meta-learning}, and a description of the data matching framework follows in Section \ref{sec:data matching}. Section \ref{sec:factorzied dd} discusses different types of factorized dataset distillation. Next, we report the performance of various distillation algorithms in Section \ref{sec:performance comparison}. Applications with dataset distillation are shown in Section \ref{sec:application}, and finally we discuss challenges and future directions in Section \ref{sec:challenges}.

\section{Background}
\label{sec:background}
Before introducing dataset distillation, we first define some notations used in this paper. For a dataset $\mathcal{T}=\{(\boldsymbol{x}_i, y_i)\}_{i=1}^{m}$, $\boldsymbol{x}_i\in \mathbb{R}^d$, $d$ is the dimension of the input data, and $y_i$ is the label. We assume that $(\boldsymbol{x}_i,y_i)$ for $1\leq i \leq m$ are independent and identically distributed (i.i.d.) random variables drawn from the data generating distribution $\mathcal{D}$. We employ $f_{\boldsymbol{\theta}}$ to denote the neural network parameterized by $\boldsymbol{\theta}$, and $f_{\boldsymbol{\theta}}(\boldsymbol{x})$ is the prediction or output of $f_{\boldsymbol{\theta}}$ at $\boldsymbol{x}$. Moreover, we also define the loss between the prediction and ground truth as $\ell(f_{\boldsymbol{\theta}}(\boldsymbol{x}),y)$, and the expected risk in terms of $\boldsymbol{\theta}$ is defined as
\begin{equation}
    \mathcal{R}_{\mathcal{D}}(\boldsymbol{\theta}) = \mathbb{E}_{(\boldsymbol{x},y)\sim \mathcal{D}}\left[\ell\left(f_{\boldsymbol{\theta}}\left(\boldsymbol{x}\right),y\right) \right].
\end{equation}
Since the data generating distribution $\mathcal{D}$ is unknown, evaluating the expected risk $\mathcal{R}_{\mathcal{D}}$ is impractical. Therefore, a practical way to estimate the expected risk is by the empirical risk $\mathcal{R}_\mathcal{T}$, which is defined as
\begin{equation}
    \mathcal{R}_{\mathcal{T}}(\boldsymbol{\theta}) = \mathbb{E}_{(\boldsymbol{x},y)\sim \mathcal{T}}\left[\ell\left(f_{\boldsymbol{\theta}}\left(\boldsymbol{x}\right),y\right) \right] = \frac{1}{m}\sum_{i=1}^{m} \ell \left(f_{\boldsymbol{\theta}}\left(\boldsymbol{x}_i\right),y_i\right).
\end{equation}

For the training algorithm $\texttt{alg}$, $\texttt{alg}(\mathcal{T}, \boldsymbol{\theta}^{(0)})$ denotes the learned parameters returned by empirical risk minimization (ERM) on the dataset $\mathcal{T}$ with the initialized parameter $\boldsymbol{\theta}^{(0)}$:
\begin{equation}
    \texttt{alg}(\mathcal{T}, \boldsymbol{\theta}^{(0)}) = \arg\min_{\boldsymbol{\theta}} \mathcal{R}_{\mathcal{T}}(\boldsymbol{\theta}).
\end{equation}
In the deep learning paradigm, gradient descent is the dominant training algorithm to train a neural network by minimizing the empirical risk step by step. Specifically, the network's parameters are initialized with $\boldsymbol{\theta}^{(0)}$, and then the parameter is iteratively updated according to the gradient of empirical risk:
\begin{equation}
\label{eq:gradient descent}
    \boldsymbol{\theta}^{(k+1)} = \boldsymbol{\theta}^{(k)} - \eta \boldsymbol{g}_{\mathcal{T}}^{(k)},
\end{equation}
where $\eta$ is the learning rate and $ \boldsymbol{g}_{\mathcal{T}}^{(k)}=\nabla_{\boldsymbol{\theta}^{(k)}} \mathcal{R}_{\mathcal{T}}(\boldsymbol{\theta}^{(k)})$ is the gradient.
%leveraging $\texttt{alg}$ on the dataset $\mathcal{T}$ with the initialized parameter $\boldsymbol{\theta}^{(0)}$. Em
We omit $\boldsymbol{\theta}^{(0)}$ and use $\texttt{alg}(\mathcal{T})$ if there is no ambiguity for the sake of simplicity. Then, the model $f$ trained on the dataset $\mathcal{T}$ can be denoted as $f_{\texttt{alg}(\mathcal{T})}$.

Because deep learning models are commonly extremely overparameterized, {\it i.e.,} the number of model parameters overwhelms the number of training examples, the empirical risk readily reaches zero. In this case, the generalization error, which measures the difference between the expected risk and the empirical risk, can be solely equal to the expected risk, which is reflected by test loss or test error in practical pipelines.

\subsection{Formalizing Dataset Distillation}

Given a target dataset (source training dataset) $\mathcal{T}=\{(\boldsymbol{x}_i,y_i)\}_{i=1}^m$, the objective of dataset distillation is to extract the knowledge of $\mathcal{T}$ into a small synthetic dataset $\mathcal{S}=\{(\boldsymbol{s}_j,y_j)\}_{j=1}^n$, where $n\ll m$, and the model trained on the small distilled dataset $\mathcal{S}$ can achieve a comparable generalization performance to the large original dataset $\mathcal{T}$:
\begin{equation}
    \mathbb{E}_{(\boldsymbol{x},y)\sim \mathcal{D} \atop \boldsymbol{\theta}^{(0)}\sim \mathbf{P}}\left[\ell\left(f_{\texttt{alg}(\mathcal{T})}\left(\boldsymbol{x}\right),y\right) \right] \simeq \mathbb{E}_{(\boldsymbol{x},y)\sim \mathcal{D} \atop \boldsymbol{\theta}^{(0)}\sim \mathbf{P} }\left[\ell\left(f_{\texttt{alg}(\mathcal{S})}\left(\boldsymbol{x}\right),y\right) \right].
\end{equation}

Because the training algorithm $\texttt{alg}\left(\mathcal{S},\boldsymbol{\theta}^{(0)}\right)$ is determined by the training set $\mathcal{S}$ and the initialized network parameter $\boldsymbol{\theta}^{(0)}$, many dataset distillation algorithms will take expectation on $\mathcal{S}$ {\it w.r.t.} $\boldsymbol{\theta}^{(0)}$ to improve the robustness of the distilled dataset $\mathcal{S}$ to different parameter initialization, and the objective function has the form of $\mathbb{E}_{ \boldsymbol{\theta}^{(0)}\sim \mathbf{P}}\left[\mathcal{L} \left(\mathcal{S} \right ) \right]$. In the following narrative, we omit this expectation {\it w.r.t.} initialization $\boldsymbol{\theta}^{(0)}$ for the sake of simplicity.

\begin{figure*}[t]
\centering

\subfigure[]{
\begin{minipage}[b]{0.35\textwidth}
\centering
    		\includegraphics[width=0.9\columnwidth]{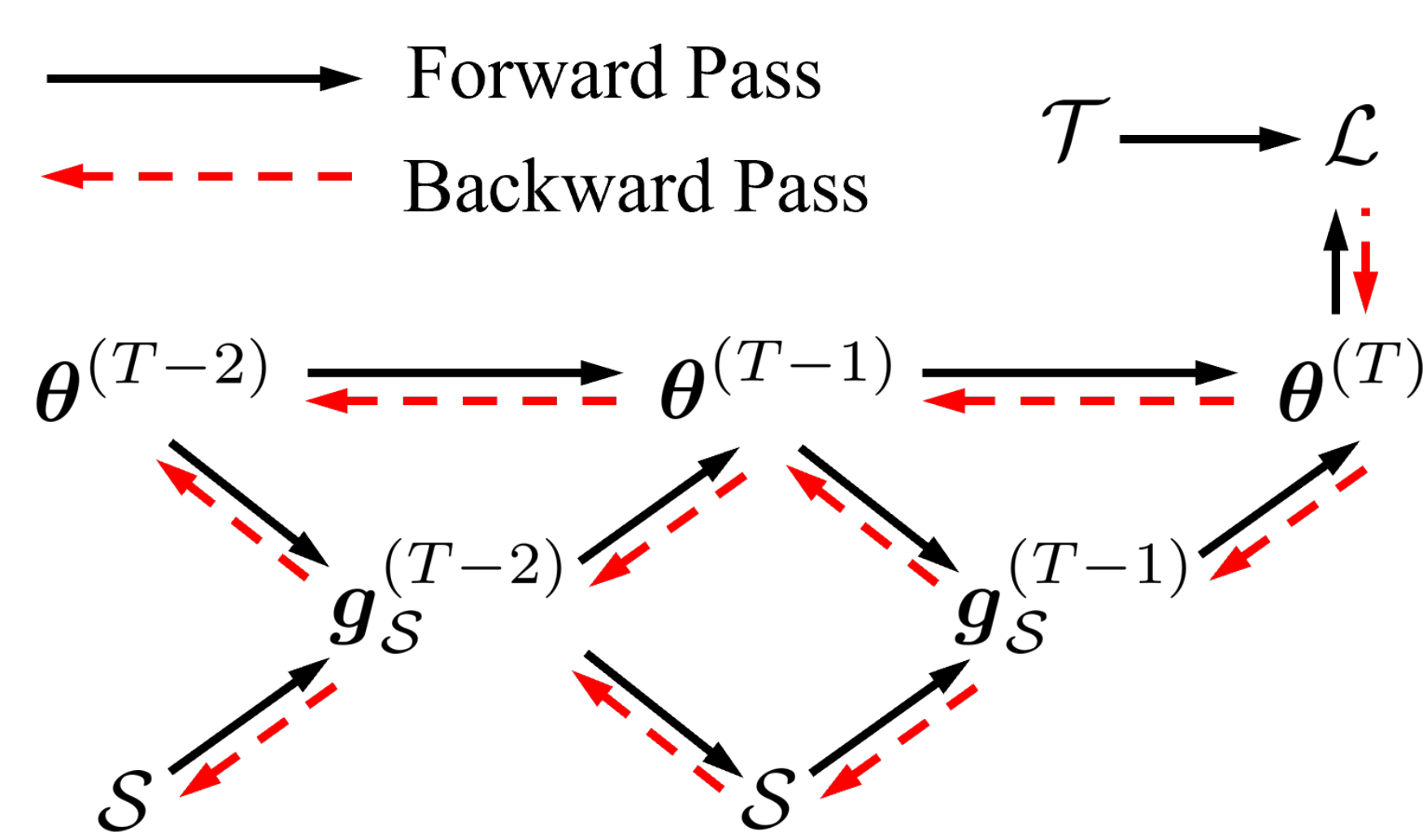}
    		\end{minipage}
		\label{figure:metaloss}   
    	}
\subfigure[]{
\begin{minipage}[b]{0.3\textwidth}
\centering
    		\includegraphics[width=0.9\columnwidth]{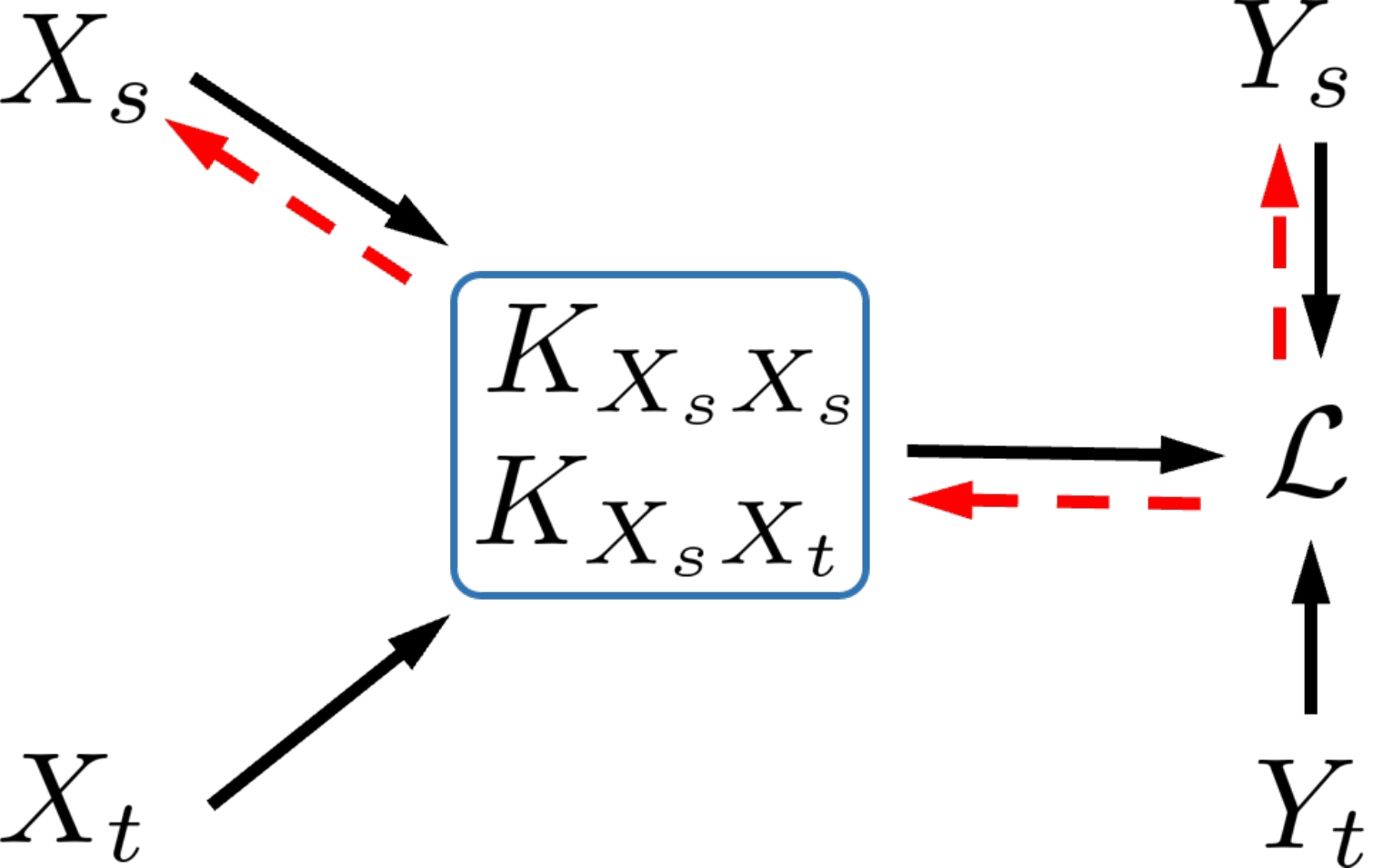}
    		\end{minipage}
		\label{figure:kernelloss}   
    	}
\subfigure[]{
\begin{minipage}[b]{0.26\textwidth}
\centering
    		\includegraphics[width=0.9\columnwidth]{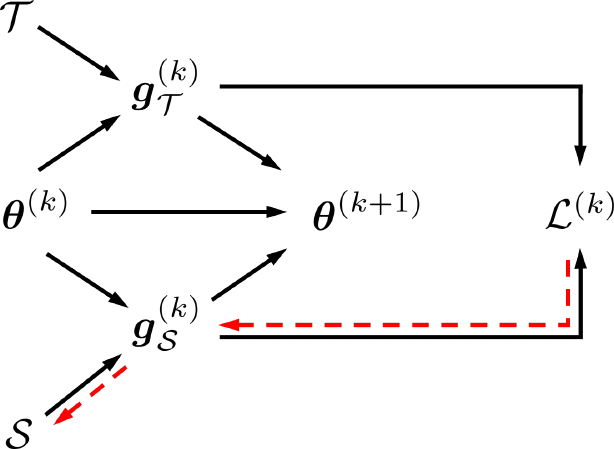}
    		\end{minipage}
		\label{figure:matchloss}
    	}
%\vspace{0.1cm}
\caption{ (a) Meta-gradient backpropagation in BPTT \citep{zhou2022dataset}; (b) Meta-gradient backpropagation in kernel ridge regression; and (c)  {Meta-gradient backpropagation in gradient matching. The gradient  $g_{\mathcal{S}}^{(T)} = \nabla_{ \boldsymbol{\theta}^{(T)}}\mathcal{R}_{\mathcal{S}}(\boldsymbol{\theta}^{(T)})$.}}
\label{figure:high dimension}
\end{figure*}

\section{Meta-Learning Framework}
\label{sec:meta-learning}
By assuming $\mathcal{R}_{\mathcal{D}}\left({\texttt{alg}(\mathcal{S})}\right) \simeq \mathcal{R}_{\mathcal{T}}\left({\texttt{alg}(\mathcal{S})}\right)$, the target dataset $\mathcal{T}$ is employed as the validation set in terms of the model ${\texttt{alg}(\mathcal{S})}$, and in consequence, the objective of DD can be converted to minimize $\mathcal{R}_{\mathcal{T}}\left({\texttt{alg}(\mathcal{S})}\right)$ to improve the generalization performance of ${\texttt{alg}(\mathcal{S})}$ in terms of $\mathcal{D}$. {To this end, dataset distillation falls into the meta-learning area: the hyperparameter $\mathcal{S}$ is updated by the meta (or outer) algorithm, and the base (or inner) algorithm solves the conventional supervised learning problem {\it w.r.t.} the synthetic dataset $\mathcal{S}$} \citep{hospedales2021meta}; and the dataset distillation task can be formulated to a bilevel optimization problem as follows:
\begin{equation}
\mathcal{S}^\ast = \arg\min_{\mathcal{S}}\mathcal{R}_{\mathcal{T}}\left({\texttt{alg}(\mathcal{S})}\right) \quad  \text{(outer loop)}
\end{equation}
subject to
\begin{equation}
\label{eq: inner loop}
    \texttt{alg}(\mathcal{S})=\arg\min _{\boldsymbol{\theta}} \mathcal{R}_\mathcal{S}\left({\boldsymbol{\theta}}\right) \quad \text{(inner loop)}
\end{equation}
The inner loop optimizes the model parameters based on the synthetic dataset and is often realized by gradient descent for neural networks or regression for kernel method. During the outer loop iteration, the synthetic set is updated by minimizing the model's risk in terms of the target dataset. With the nested loop, the synthetic dataset gradually converges to one of the optima. 

According to the model and optimization methods used in the inner loop, the meta-learning framework of DD can be further classified into two sub-categories of backpropagation through time time (BPTT) approach and kernel ridge regression (KRR) approach.

\subsection{Backpropagation Through Time Approach}

As shown in the above formulation of bilevel optimization, the objective function of DD can be directly defined as the meta-loss of
\begin{equation}
    \mathcal{L}(\mathcal{S}) = \mathcal{R}_{\mathcal{T}}\left(\texttt{alg}\left(\mathcal{S}\right)\right).
\end{equation}
Then the distilled dataset is updated by $\mathcal{S} = \mathcal{S} - \alpha \nabla_\mathcal{S} \mathcal{L}(\mathcal{S})$ with the step size $\alpha$. However, in the inner loop of Eq. \ref{eq: inner loop}, the neural network is trained by iterative gradient descent (Eq. \ref{eq:gradient descent}), which yields a series of intermediate parameter states of $\{\boldsymbol{\mathcal{\theta}}^{(0)}, \boldsymbol{\mathcal{\theta}}^{(1)}, \cdots, \boldsymbol{\mathcal{\theta}}^{(T)} \}$, backpropagation through time \citep{werbos1990backpropagation} is required to recursively compute the meta-gradient $\nabla_\mathcal{S}\mathcal{L}(\mathcal{S})$: 
\begin{equation}
    \nabla_\mathcal{S}\mathcal{L}(\mathcal{S}) = \frac{\partial \mathcal{L}}{\partial \mathcal{S}} = \frac{\partial \mathcal{L}}{\partial \boldsymbol{\theta}^{(T)}}\left( \sum_{k=0}^{k=T} \frac{\partial \boldsymbol{\theta}^{(T)}}{\partial \boldsymbol{\theta}^{(k)}} \cdot \frac{\partial \boldsymbol{\theta}^{(k)}}{\partial \mathcal{S}} \right), 
\end{equation}
and
\begin{equation}
    \frac{\partial \boldsymbol{\theta}^{(T)}}{\partial \boldsymbol{\theta}^{(k)}} = \prod_{i=k+1}^{T} \frac{\partial \boldsymbol{\theta}^{(i)}}{\partial \boldsymbol{\theta}^{(i-1)}}.
\end{equation}
We present the meta-gradient backpropagation of BPTT in Figure \ref{figure:metaloss}. Due to the requirement of unrolling the recursive computation graph, BPTT is both computationally expensive and memory demanding, which also severely affects the final distillation performance.

To alleviate the inefficiency in unrolling the long parameter path of $\{\boldsymbol{\mathcal{\theta}}^{(0)}, \cdots, \boldsymbol{\mathcal{\theta}}^{(T)} \}$, \citet{wang2018dataset} adopted a single-step optimization {\it w.r.t.} the model parameter from $\boldsymbol{\theta}^{(0)}$ to $\boldsymbol{\theta}^{(1)}$, and the meta-loss was computed based on $\boldsymbol{\theta}^{(1)}$ and the target dataset $\mathcal{T}$: {
\begin{equation}
    \boldsymbol{\theta}^{(1)} = \boldsymbol{\theta}^{(0)} - \eta \nabla_{ \boldsymbol{\theta}^{(0)}}\mathcal{R}_{\mathcal{S}}(\boldsymbol{\theta}^{(0)}) \quad \text{and} \quad \mathcal{L} = \mathcal{R}_{\mathcal{T}}(\boldsymbol{\theta}^{(1)}).
\end{equation}
Therefore, the distilled data $\mathcal{S}$ and learning rate $\eta$ can be efficiently updated via the short-range BPTT as follows:
\begin{equation}
    \mathcal{S} =  \mathcal{S} - \alpha_{\boldsymbol{s}_i } \nabla \mathcal{L} \quad \text{and} \quad \eta = \eta - \alpha_\eta \nabla \mathcal{L}.
\end{equation}}
Unlike freezing distilled labels, \citet{sucholutsky2021soft} extended the work of \citet{wang2018dataset} by learning a soft-label in the synthetic dataset $\mathcal{S}$: {\it i.e.}, the label $y$ in the synthetic dataset is also trainable for better information compression, {\it i.e.}, $y_i = y_i - \alpha \nabla_{y_i} \mathcal{L}$ for $(\boldsymbol{s}_i, y_i)\in \mathcal{S}$. Similarly, \citet{bohdal2020flexible} also extended the standard example distillation to label distillation by solely optimizing the labels of synthetic datasets. Moreover, these researchers provided improvements on the efficiency of long inner loop optimization via (1) iteratively updating the model parameters $\boldsymbol{\theta}$ and the distilled labels $y$, {{\it i.e.}, one outer step followed by only one inner step for faster convergence}; and (2) fixing the feature extractor of neural networks and solely updating the last linear layer with ridge regression to avoid second-order meta-gradient computation. 
%employing ridge regression to update the solution of the last linear layer of networks to avoid second-order gradient computation. 
Although the BPTT framework has been shown to underperform other algorithms, \citet{deng2022remember} empirically demonstrated that adding momentum term and longer unrolled trajectory ($200$ steps) in the inner loop optimization can considerably enhance the distillation performance, and the inner loop of model training becomes
\begin{equation}
    \boldsymbol{\theta}^{(k+1)} = \boldsymbol{\theta}^{(k)} - \eta \boldsymbol{m}^{(k+1)},
\end{equation}
where
\begin{equation}
    \boldsymbol{m}^{(k+1)} = \beta \boldsymbol{m}^{(k)} +  \nabla_{\boldsymbol{\theta}^{(k)}} \mathcal{R}_{\mathcal{T}}(\boldsymbol{\theta}^{(k)}) \quad \text{s.t.} \quad \boldsymbol{m}^{(0)} = \boldsymbol{0}.
\end{equation}

\subsection{Kernel Ridge Regression Approach}
Although multistep gradient descent can gradually approach the optimal network parameters in terms of the synthetic dataset during the inner loop, this iterative algorithm makes the meta-gradient backpropagation highly inefficient, as shown in BPTT. 
%Because the solutions of deep non-linear neural networks are commonly intractable, multi-step gradient descent is used to get the optimal parameters in terms of the synthetic dataset. Nevertheless, the iterative algorithm makes the meat-gradient backpropagation very inefficient, as shown in BPTT. 
Considering the existence of closed-form solutions in the kernel regression regime, \citet{nguyen2020dataset} replaced the neural network in the inner loop with a kernel model, which bypasses the recursive backpropagation of the meta-gradient. For the regression model $f(\boldsymbol{x})=\boldsymbol{w}^\top\psi(\boldsymbol{x})$, where $\psi(\cdot)$ is a nonlinear mapping and the corresponding kernel is $K(\boldsymbol{x},\boldsymbol{x}^\prime)=\langle \psi(\boldsymbol{x}), \psi(\boldsymbol{x}^\prime)\rangle$, there exists a closed-form solution for $\boldsymbol{w}$ when the regression model is trained on $\mathcal{S}$ with kernel ridge regression (KRR):
\begin{equation}
   \boldsymbol{w} =  \psi(X_{s})^\top\left(\mathbf{K}_{X_s X_s}+\lambda I\right)^{-1}y_{s},
\end{equation}
where $\mathbf{K}_{X_s X_s} = [K(\boldsymbol{s}_i,\boldsymbol{s}_j)]_{ij}\in \mathbb{R}^{n\times n}$ is called the {\it kernel matrix} or {\it Gram matrix} associated with $K$ and the dataset $\mathcal{S}$, and $\lambda > 0$ is a fixed regularization parameter \citep{petersen2008matrix}. Therefore, the mean square error (MSE) of predicting $\mathcal{T}$ with the model trained on $\mathcal{S}$ is
\begin{equation}
\label{eq:mse}
    \mathcal{L}(\mathcal{S}) = \frac{1}{2}\left\|y_t-\mathbf{K}_{X_t X_s}\left(\mathbf{K}_{X_s X_s}+\lambda I\right)^{-1} y_s\right\|^2,
\end{equation}
where $\mathbf{K}_{X_t X_s}=[K(\boldsymbol{x}_i,\boldsymbol{s}_j)]_{ij}\in \mathbb{R}^{m\times n}$. Then the distilled dataset is updated via the meta-gradient of the above loss. Due to the closed-form solution in KRR, $\boldsymbol{\theta}$ does not require an iterative update and the backward pass of the gradient thus bypasses the recursive computation graph, as shown in Figure \ref{figure:kernelloss}. 

In the KRR regime, the synthetic dataset $\mathcal{S}$ can be directly updated by backpropagating the meta-gradient through the kernel function. Although this formulation is solid, this algorithm is designed in the KRR scenario and only employs simple kernels, which causes performance drops when the distilled dataset is transferred to train neural networks. \citet{jacot2018neural} proposed the neural tangent kernel (NTK) theory that proves the equivalence between training infinite-width neural networks and kernel regression. With this equivalence, \citet{nguyen2021dataset} employed infinite-width networks as the kernel for dataset distillation, which narrows the gap between the scenarios of KRR and deep learning. 

However, every entry in the kernel matrix must be calculated separately via the kernel function, and thus computing the kernel matrix $\mathbf{K}_{X_s X_t}$ has the time complexity of $\mathcal{O}(|\mathcal{T}||\mathcal{S}|)$, which is severely inefficient for large-scale datasets with huge $|\mathcal{T}|$. To tackle this problem, \citet{loo2022efficient} replaced the NTK kernel with neural network Gaussian process (NNGP) kernel that only considers the training dynamic of the last-layer classifier for speed up. With this replacement, the random features $\psi(\boldsymbol{x})$ and $\psi(\boldsymbol{s})$ can be explicitly computed via multiple sampling from the Gaussian process, and thus the kernel matrix computation can be decomposed into random feature calculation and random feature matrix multiplication. Because matrix multiplication requires negligible amounts of time for small distilled datasets, the time complexity of kernel matrix computation degrades to $\mathcal{O}(|\mathcal{T}|+|\mathcal{S}|)$. In addition, these authors demonstrated two issues of MSE loss (Eq. \ref{eq:mse}) used in \citep{nguyen2020dataset} and \citep{nguyen2021dataset}: (1) over-influence on corrected data: the correctly classified examples in $\mathcal{T}$ can induce larger loss than the incorrectly classified examples; and (2) unclear probabilistic interpretation for classification tasks. To overcome these issues, they propose to apply a cross-entropy (CE) loss with Platt scaling \citep{platt1999probabilistic} to replace the MSE loss:
\begin{equation}
    \mathcal{L}_{\tau} = \text{CE}(y_t, \hat{y}_t / \tau),
\end{equation}
where $\tau$ is a positive learned temperature scaling parameter, and the prediction $\hat{y}_t = \mathbf{K}_{X_t X_s}\left(\mathbf{K}_{X_s X_s}+\lambda I\right)^{-1} y_s$ is still calculated using the KRR formula.

A similar efficient method was also proposed by \citet{zhou2022dataset}, which also focused on solving the last-layer classifier in neural networks with KRR. Specifically, the neural network $f_{\boldsymbol{\theta}} = g_{\boldsymbol{\theta}_2} \circ h_{\boldsymbol{\theta}_1}$ can be decomposed with the feature extractor $h$ and the linear classifier $g$. Then these coworkers fixed the feature extractor $h$ and the linear classifier $g$ possesses a closed-form solution with KRR, and the distilled data are accordingly optimized with the MSE loss:
\begin{equation}
    \mathcal{L}(\mathcal{S}) = \frac{1}{2}\left\|y_t-\mathbf{K}_{X_t X_s}^{\boldsymbol{\theta}_1^{(k)}}\left(\mathbf{K}_{X_s X_s}^{\boldsymbol{\theta}_1^{(k)}}+\lambda I\right)^{-1} y_s\right\|^2
\end{equation}
and
\begin{equation}
    \boldsymbol{\theta}_1^{(k)} = \boldsymbol{\theta}_1^{(k-1)} - \eta \nabla_{ \boldsymbol{\theta}_1^{(k-1)}}\mathcal{R}_{\mathcal{S}}(\boldsymbol{\theta}_1^{(k-1)}),
\end{equation}
where the kernel matrices $\mathbf{K}_{X_t X_s}^{\boldsymbol{\theta}_1}$ and $\mathbf{K}_{X_s X_s}^{\boldsymbol{\theta}_1}$ are induced by $h_{\boldsymbol{\theta}_1}(\cdot)$. Notably, although the feature extractor $h_{\boldsymbol{\theta}_1}$ is continuously updated, the classifier $g_{\boldsymbol{\theta}_2}$ is directly solved by KRR, and thus the meta-gradient backpropagation side-steps the recursive computation graph.

\subsection{Discussion}

From the loss surface perspective \citep{li2018visualizing}, the meta-learning framework of minimizing $\mathcal{R}_{\mathcal{T}}\left(\texttt{alg}\left(\mathcal{S}\right)\right)$ can be considered to mimic the local minima of target data with the distilled data. However, the loss landscape {\it w.r.t.} parameters is closely related to the network architecture, while only one type of small network is used in the BPTT approach. Consequently, there is a moderate performance drop when the distilled dataset is employed to train other complicated networks. Moreover, a long unrolled trajectory and second-order gradient computation are also two key challenges for BPTT approach, which hinder its efficiency. The KRR approach compensates for these shortcomings by replacing networks with the nonparametric kernel model which admits closed-form solution. Although KRR is nonparametric and does not involve neural networks during the distillation process, previous research has shown that the training dynamic of neural networks is equal to the kernel method when the width of networks becomes infinite \citep{jacot2018neural,golikov2022neural,bietti2019inductive,bietti2019inductive}, which partially guarantees the feasibility of the kernel regression approach and explains its decent performance when transferred to wide neural networks.

\section{Data Matching Framework}
\label{sec:data matching}

Although it is not feasible to explicitly extract information from target data and then inject them into synthetic data, information distillation can be achieved by implicitly aligning the byproducts of target data and synthetic data from different aspects; and this byproduct matching allows synthetic data to imitate the influence of target data on model training. The objective function of data matching can be summarized as follows.
\begin{equation}
    \mathcal{L}(\mathcal{S}) = \sum_{k=0}^{T} D\left( \phi(\mathcal{S},\boldsymbol{\theta}^{(k)}), \phi(\mathcal{T},\boldsymbol{\theta}^{(k)}) \right)
\end{equation}
subject to
\begin{equation}
\label{eq:dm_update}
    \boldsymbol{\theta}^{(k)} = \boldsymbol{\theta}^{(k-1)} - \eta \nabla_{\boldsymbol{\theta}^{(k-1)}} \mathcal{R}_{\mathcal{S}}\left(\boldsymbol{\theta}^{(k-1)}\right),
\end{equation}
where $D(\cdot,\cdot)$ is a distance function, and $\phi(\cdot)$ maps the dataset $\mathcal{S}$ or $\mathcal{T}$ to other informative spaces, such as gradient, parameter, and feature spaces. In practical implementation, the full datasets of $\mathcal{S}$ and $\mathcal{T}$ are often replaced with randomly sampled batches of $\mathcal{B}_\mathcal{S}$ and $\mathcal{B}_\mathcal{T}$ for memory saving and faster convergence. 

Compared to the aforementioned meta-learning framework, the data matching loss not only focuses on the final parameter $\texttt{alg}(\mathcal{S})$ but also supervises the intermediate states, as shown in the sum operation $\sum_{k=0}^{T}$. By this, the distilled data can better imitate the influence of target data on training networks at different training stages.

\begin{figure*}[t]
\centering

\subfigure[Trajectory matching]{
\begin{minipage}[b]{0.62\textwidth}
\centering
    		\includegraphics[width=0.9\columnwidth]{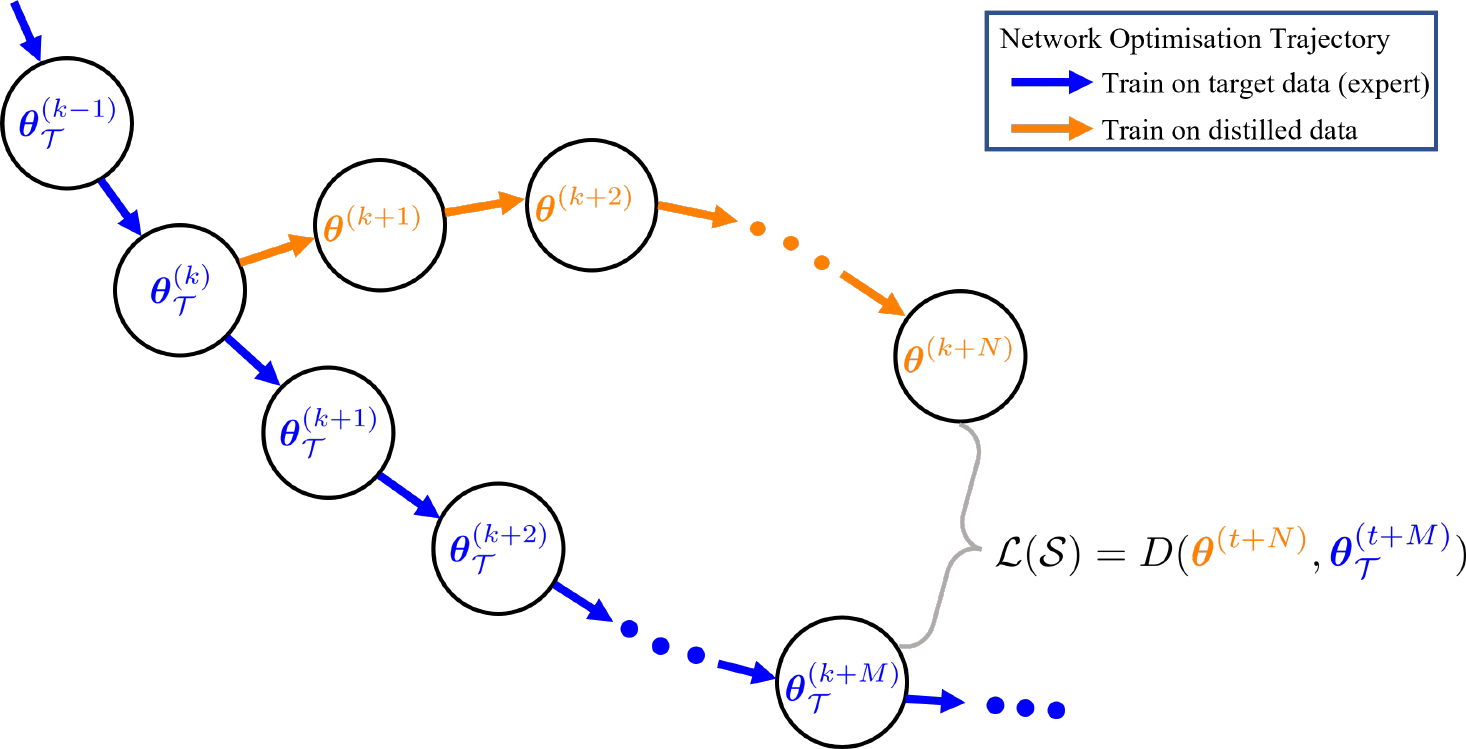}
    		\end{minipage}
		\label{figure:mtt_illustration}   
    	}
\subfigure[Distribution matching]{
\begin{minipage}[b]{0.33\textwidth}
\centering
    		\includegraphics[width=0.9\columnwidth]{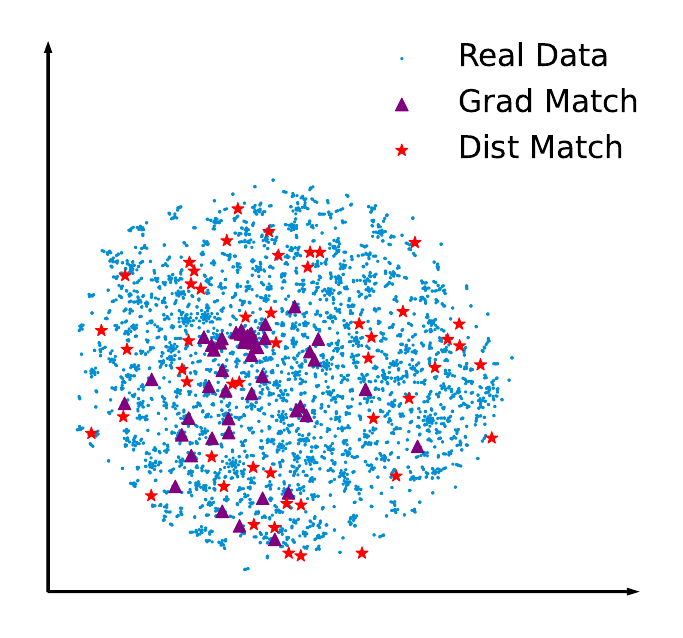}
    		\end{minipage}
		\label{figure:dm_illustration}   
    	}
\caption{(a) An illustration of trajectory matching \citep{cazenavette2022distillation}; (b)  {T-SNE \citep{van2008visualizing} visualization on the first class of CIFAR-10. Compared to gradient matching (Grad Match, marked with purple triangles), distribution matching (Dist Match, marked with red stars) can more comprehensively cover the real data distribution in feature space.}}
\label{figure:match illustration}
\end{figure*}

\subsection{Gradient Matching Approach}
To achieve comparable generalization performance, a natural approach is to imitate the model parameters, {\it i.e.}, matching the training trajectories introduced by $\mathcal{S}$ and $\mathcal{T}$. With a fixed parameter initialization, the training trajectory of $\{\boldsymbol{\mathcal{\theta}}^{(0)}, \boldsymbol{\mathcal{\theta}}^{(1)}, \cdots, \boldsymbol{\mathcal{\theta}}^{(T)} \}$ is equal to a series of gradients $\{\boldsymbol{g}^{(0)}, \cdots, \boldsymbol{g}^{(T)}\}$. Therefore, matching the gradients induced by $\mathcal{S}$ and $\mathcal{T}$ is a convincing proxy to mimic the influence on model parameters \citep{zhao2021dataset}, and the objective function is formulated as
\begin{equation}
    \mathcal{L}(\mathcal{S}) =  \sum_{k=0}^{T-1} D\left( \nabla_{\boldsymbol{\theta}^{(k)}} \mathcal{R}_{\mathcal{S}}\left(\boldsymbol{\theta}^{(k)}\right), \nabla_{\boldsymbol{\theta}^{(k)}} \mathcal{R}_{\mathcal{T}}\left(\boldsymbol{\theta}^{(k)}\right) \right).
\end{equation}

The difference $D$ between gradients is measured in the layer-wise aspect:
\begin{equation}
    D\left( \boldsymbol{g}_{\mathcal{S}}, \boldsymbol{g}_{\mathcal{T}} \right) = \sum_{l=1}^L \texttt{dis}\left( \boldsymbol{g}_{\mathcal{S}}^{l}, \boldsymbol{g}_{\mathcal{T}}^{l} \right),
\end{equation}
where $\boldsymbol{g}^{l}$ denotes the gradient of $i$-th layer, and $\texttt{dis}$ is the sum of cosine distance as follows:
\begin{equation}
\label{eq:dis}
    \texttt{dis}\left(\boldsymbol{A}, \boldsymbol{B}\right) = \sum_{i=1}^{\texttt{out}} \left(1 - \frac{\boldsymbol{A}_i \boldsymbol{B}_i}{ \| \boldsymbol{A}_i \| \| \boldsymbol{B}_i \|} \right),
\end{equation}
where $\texttt{out}$ denotes the number of output channels for specific layer; and $\boldsymbol{A}_i$ and $\boldsymbol{B}_i$ are the flatten gradients in the $i$-th channel.

To improve the convergence speed in practical implementation, \citet{zhao2021dataset} proposed to match gradient in class-wise as follows:
\begin{equation}
\label{eq:interclass_gm}
    \mathcal{L}^{(k)}=\sum_{c=1}^C D\left(\nabla_{\boldsymbol{\theta}^{(k)}} \mathcal{R}\left(\mathcal{B}_\mathcal{S}^c, \boldsymbol{\theta}^{(k)}\right), \nabla_{\boldsymbol{\theta}^{(k)}} \mathcal{R}\left(\mathcal{B}_\mathcal{T}^c,  \boldsymbol{\theta}^{(k)}\right) \right),
\end{equation}
where $c$ is the class index and $\mathcal{B}_\mathcal{S}^c$ and $\mathcal{B}_\mathcal{T}^c$ denotes the batch examples belong to the $c$-th class. However, according to \citep{lee2022dataset}, the class-wise gradient matching pays much attention to the class-common features and overlooks the class-discriminative features in the target dataset, and the distilled synthetic dataset $\mathcal{S}$ does not possess enough class-discriminative information, especially when the target dataset is fine-grained, {\it i.e.}, class-common features are dominant. Based on this finding, these coworkers proposed an improved objective function of
\begin{equation}
\label{eq:intraclass_gm}
    \mathcal{L}^{(k)}=D\left(\sum_{c=1}^C\nabla_{\boldsymbol{\theta}^{(k)}} \mathcal{R}\left(\mathcal{B}_\mathcal{S}^c, \boldsymbol{\theta}^{(k)}\right), \sum_{c=1}^C \nabla_{\boldsymbol{\theta}^{(k)}} \mathcal{R}\left(\mathcal{B}_\mathcal{T}^c, \boldsymbol{\theta}^{(k)}\right) \right)
\end{equation}
to better capture contrastive signals between different classes through the sum of loss gradients between classes. A similar approach was proposed by \citet{jiang2022delving}, which employed both intraclass and interclass gradient matching by combining Eq. \ref{eq:interclass_gm} and Eq. \ref{eq:intraclass_gm}. Moreover, these researchers measure the difference of gradients by considering the magnitude instead of only considering the angle, {\it i.e.}, the cosine distance, and the distance function of Eq. \ref{eq:dis} is improved to
\begin{equation}
    \texttt{dis}\left(\boldsymbol{A}, \boldsymbol{B}\right) = \sum_{i=1}^{\texttt{out}} \left(1 - \frac{\boldsymbol{A}_i \boldsymbol{B}_i}{ \| \boldsymbol{A}_i \| \| \boldsymbol{B}_i \|} + \| \boldsymbol{A}_i - \boldsymbol{B}_i \|\right).
\end{equation}
To alleviate easy overfitting on the small dataset $\mathcal{S}$, \citet{kim2022dataset} proposed to perform inner loop optimization on the target dataset $\mathcal{T}$ instead of $\mathcal{S}$, {\it i.e.}, replaced the parameter update of Eq. \ref{eq:dm_update} with
\begin{equation}
    \boldsymbol{\theta}^{(k+1)} = \boldsymbol{\theta}^{(k)} - \eta \nabla_{\boldsymbol{\theta}^{(k)}} \mathcal{R}_\mathcal{T}\left(\boldsymbol{\theta}^{(k)}\right).
\end{equation}

Although data augmentation facilitates a large performance increase in conventional network training, conducting augmentation on distilled datasets demonstrates no improvement on the final test accuracy, because the characteristics of synthetic images are different from those of natural images and are not optimized under the supervision of various transformations. To leverage data augmentation on synthetic datasets, \citet{zhao2021dsa} designed data Siamese augmentation (DSA) that homologously augments the distilled data and the target data during the distillation process as follows:
\begin{equation}
    \mathcal{L} = D\left(\nabla_{\boldsymbol{\theta}} \mathcal{R}\left(\mathcal{A}_{\omega}\left(\mathcal{B}_\mathcal{S} \right), \boldsymbol{\theta}\right), \nabla_{\boldsymbol{\theta}} \mathcal{R}\left(\mathcal{A}_{\omega}\left(\mathcal{B}_\mathcal{T} \right), \boldsymbol{\theta}\right) \right),
\end{equation}
where $\mathcal{A}$ is a family of image transformations such as cropping and flipping that are parameterized with $\omega^{\mathcal{S}}$ and $\omega^{\mathcal{T}}$ for synthetic and target data, respectively. In DSA, the augmented form of distilled data has a consistent correspondence {\it w.r.t.} the augmented form of the target data, {\it i.e.}, $\omega^{\mathcal{S}} = \omega^{\mathcal{T}} = \omega$; and $\omega$ is randomly picked from $\mathcal{A}$ at different iterations. Notably, the transformation $\mathcal{A}$ requires to be differentiable {\it w.r.t.} the synthetic dataset $\mathcal{S}$ for backpropagation:
\begin{equation}
\frac{\partial D(\cdot)}{\partial \mathcal{S}}=\frac{\partial D(\cdot)}{\partial \nabla_\theta \mathcal{L}(\cdot)} \frac{\partial \nabla_\theta \mathcal{L}(\cdot)}{\partial \mathcal{A}(\cdot)} \frac{\partial \mathcal{A}(\cdot)}{\partial \mathcal{S}}.
\end{equation}

Through setting $\omega^{\mathcal{S}} = \omega^{\mathcal{T}}$, DSA permits the knowledge transfer from the transformation of target images to the corresponding transformation of synthetic images. Consequently, the augmented synthetic images also possess meaningful characteristics of the natural images. Due to its superior compatibility, DSA has been widely used in many data matching methods.

\subsection{Trajectory Matching Approach}

Unlike circuitously matching the gradients, \citet{cazenavette2022distillation} directly matched the long-range training trajectory between the target dataset and the synthetic dataset. Specifically, they train models on the target dataset $\mathcal{T}$ and collect the expert training trajectory into a buffer in advance. Then the ingredients in the buffer are randomly selected to initialize the networks for training $\mathcal{S}$. After collecting the trajectories of $\mathcal{S}$, the synthetic dataset is updated by matching their parameters, as shown in Figure \ref{figure:mtt_illustration}. The objective loss of trajectory matching is defined as 
\begin{equation}
\label{eq:mtt_loss}
    \mathcal{L}= \frac{\| \boldsymbol{\theta}^{(k+N)} - \boldsymbol{\theta}_{\mathcal{T}}^{(k+M)} \|_2 ^2}{ \|\boldsymbol{\theta}_{\mathcal{T}}^{(k)} - \boldsymbol{\theta}_{\mathcal{T}}^{(k+M)} \|_2 ^2},
\end{equation}
where $\boldsymbol{\theta}_{\mathcal{T}}$ denotes the target parameter by training the model on $\mathcal{T}$, which is stored in the buffer, and $\boldsymbol{\theta}^{(k+N)}$ are the parameter obtained by training the model on $\mathcal{S}$ for $N$ epochs with the initialization of $\boldsymbol{\theta}_{\mathcal{T}}^{(k)}$. The denominator in the loss function is for normalization.

Although trajectory matching demonstrated empirical success, \citet{du2022minimizing} argued that $\boldsymbol{\theta}^{(k)}$ is generally induced by training on $\mathcal{S}$ for $k$ epochs during the natural learning process, while it was directly initialized with $\boldsymbol{\theta}_{\mathcal{T}}^{(k)}$ in trajectory matching, which causes the {\it accumulated trajectory error} that measures the difference between the parameters from real trajectories of $\mathcal{S}$ and $\mathcal{T}$. Because the accumulated trajectory error is induced by the mismatch between $\boldsymbol{\theta}_{\mathcal{T}}^{(k)}$ and real $\boldsymbol{\theta}^{(k)}$ from the natural training, these researchers alleviate this error by adding random noise to when initializing $\boldsymbol{\theta}^{(k)}$ to improve robustness {\it w.r.t.} $\boldsymbol{\theta}_{\mathcal{T}}^{(k)}$.

Compared to matching gradients, while trajectory matching side-steps second-order gradient computation, unfortunately, unrolling $N$ SGD updates are required during meta-gradient backpropagation due to the existence of $\boldsymbol{\theta}^{(k+N)}$ in Eq. \ref{eq:mtt_loss}. The unrolled gradient computation significantly increases the memory burden and impedes scalability. By disentangling the meta-gradient {\it w.r.t.} synthetic examples into two passes, \citet{cui2023scaling} greatly reduced the memory required by trajectory matching and successfully scaled trajectory matching approach to the large ImageNet-1K dataset \citep{russakovsky2015imagenet}. Motivated by knowledge distillation \citep{gou2021knowledge}, these scholars also proposed assigning soft-labels to synthetic examples with pretrained models in the buffer; and the soft-labels help learn intraclass information and consequently improve distillation performance.

\subsection{Distribution Matching Approach}
\begin{figure}[t]
    \centering
    \includegraphics[width=0.95\columnwidth]{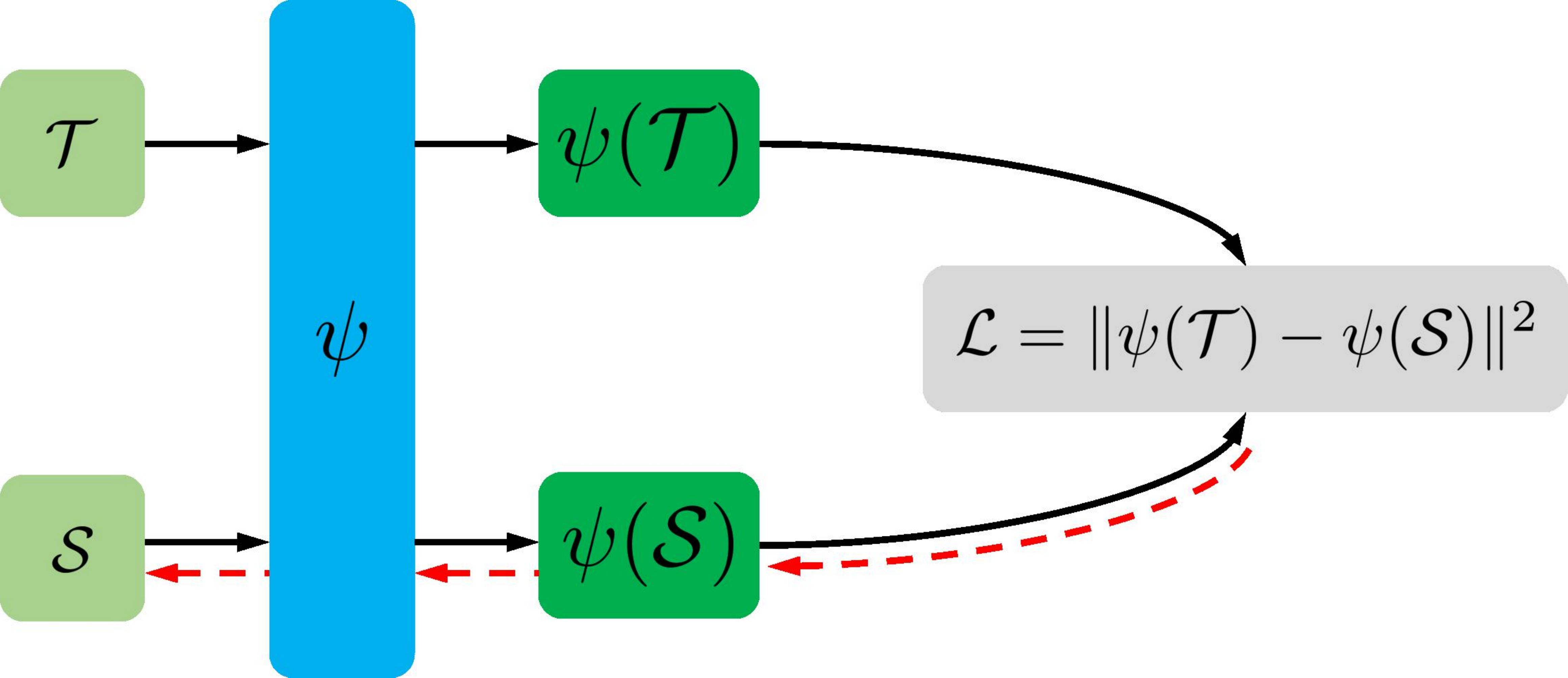}
    \caption{An illustration of distribution matching. $\psi$ is the feature extractor, and red dashed lines denote the backpropagation of gradients.}
    \label{fig:dm}
\end{figure}
Although the above parameter-wise matching shows satisfying performance, \citet{zhao2023distribution} visualized the distilled data in two-dimensional plane and revealed that there is a large distribution discrepancy between the distilled data and the target data. In other words, the distilled dataset cannot comprehensively cover the data distribution in feature space, as shown in Figure \ref{figure:dm_illustration}. Based on this discovery, these researchers proposed to match the synthetic and target data from the distribution perspective for dataset distillation. Specifically, they employed the pretrained feature extractor $\psi_{\boldsymbol{v}}$ with the parameter $\boldsymbol{v}$ to achieve mapping from the input space to the feature space. The synthetic data are optimized according to the objective function 
\begin{equation}
\label{eq:dm}
    \mathcal{L}(\mathcal{S})=\sum_{c=1}^C \|\psi_{\boldsymbol{v}}(\mathcal{B}_\mathcal{S}^c) - \psi_{\boldsymbol{v}}(\mathcal{B}_\mathcal{T}^c)  \|^2,
\end{equation}
where $c$ denotes the class index. An illustration of distribution matching is also presented in Figure \ref{fig:dm}. As shown in Eq. \ref{eq:dm}, distribution matching does not rely on the model parameters and drops bilevel optimization for less memory requirement, whereas it empirically underperforms the above gradient and trajectory matching approaches. 

\citet{wang2022cafe} improved the distribution matching from several aspects: (1) using multiple-layer features other than only the last-layer features for matching, and the feature matching loss is formulated as follows:
\begin{equation}
    \mathcal{L}_{\text{f}} = \sum_{c=1}^C \sum_{l=1}^L |f_{\boldsymbol{\theta}}^l (\mathcal{B}_\mathcal{S}^c) - f_{\boldsymbol{\theta}}^l (\mathcal{T}_\mathcal{S}^c)|^2,
\end{equation}
where $f_{\boldsymbol{\theta}}^l$ denotes the $l$-th layer features and $L$ is the total number of network layers; (2) recalling the bilevel optimization that updates $\mathcal{S}$ with different model parameters by inserting the inner loop of $\boldsymbol{\theta}^{(k+1)} = \boldsymbol{\theta}^{(k)} -  \eta\nabla_{\boldsymbol{\theta}^{(k)}} \mathcal{R}_{\mathcal{S}}(\boldsymbol{\theta}^{(k)})$;
and (3) proposing the discrimination loss in the last-layer feature space to enlarge the class distinction of synthetic data. Specifically, the synthetic feature center $\bar{f}^\mathcal{S}_{c,L}$ of each category $c$ is obtained by averaging the batch $\mathcal{B}_\mathcal{S}^c$. The objective of discrimination loss is to improve the classification ability of the feature center $\bar{\boldsymbol{F}}_{L}^{\mathcal{S}} = [\bar{f}^\mathcal{S}_{1,L}, \bar{f}^\mathcal{S}_{2,L}, \cdots, \bar{f}^\mathcal{S}_{C,L}]$ on predicting the real data feature $\boldsymbol{F}_L^{\mathcal{T}}=[{f}^\mathcal{T}_{1,L}, {f}^\mathcal{T}_{1,L}, \cdots, {f}^\mathcal{T}_{C,L}]$. The logits are first calculated by
\begin{equation}
    \mathbf{O} = \left\langle \mathbf{\boldsymbol{F}}_{L}^{\mathcal{T}}, \left(\bar{\boldsymbol{F}}_{L}^\mathcal{S}\right)^\top \right \rangle,
\end{equation}
where $\mathbf{O}\in \mathbb{R}^{N\times C}$ contains the logits of $N=C\times |\mathcal{B}_\mathcal{T}^c|$ target data points. Then the probability $p_i$ in terms of the ground-truth label is derived via $p_i = \text{softmax}(\mathbf{O}_i)$; and the classification loss is
\begin{equation}
    \mathcal{L}_{\text{d}} = \frac{1}{N} \sum_{i=1}^N \log p_i.
\end{equation}
Therefore, the total loss in \citep{wang2022cafe} for dataset distillation is $\mathcal{L}_{\text{total}} = \mathcal{L}_{\text{f}} + \beta \mathcal{L}_{\text{d}}$, where $\beta$ is a positive factor for balancing the feature matching and discrimination loss. Similar to the discrimination loss, \citet{zhao2023improved} add a classification loss $\mathcal{L}_{CE}$ as regularization to mitigate less classified synthetic data caused by the first-order moment mean matching:
\begin{equation}
    \mathcal{L}_{CE}(\mathcal{S}) = \sum_{i=1}^{|\mathcal{S}|}CE(f_{\boldsymbol{\theta}}(s_i), y_i),
\end{equation}
where the network parameter $\boldsymbol{\theta}$ is varied in different loops.

\subsection{Discussion}

The gradient matching approach can be considered to be short-range parameter matching, while its backpropagation requires second-order gradient computation. Although the trajectory matching approach makes up for this imperfection, the long-range trajectory introduces a recursive computation graph during meta-gradient backpropagation. Different from matching in parameter space, distribution matching approach employs the feature space as the match proxy and also bypasses second-order gradient computation. Although distribution matching has advantages in scalability {\it w.r.t.} high-dimensional data, it empirically underperforms trajectory matching, which might be attributed to the mismatch between the comprehensive distribution and decent distillation performance. In detail, distribution matching achieves DD by mimicking features of the target data, so that the distilled data are evenly distributed in the feature space. However, not all features are equally important and distribution matching might waste the budget on imitating less informative features, thereby undermining the distillation performance. 

\begin{figure}[t]
    \centering
    \includegraphics[width=0.98\columnwidth]{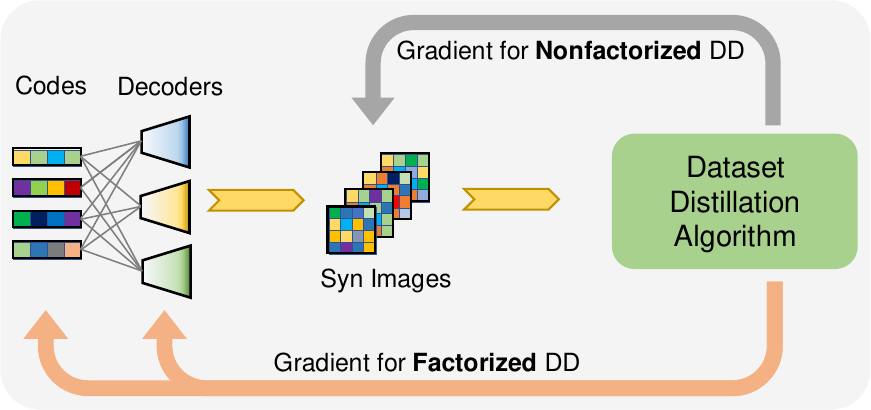}
    \caption{ {Schematic diagrams of nonfactorized dataset distillation and factorized dataset distillation. Unlike directly updating synthetic (Syn) images in nonfactorized DD, factorized DD optimizes the codes and decoders, which are then combined to generate synthetic data.}}
    \label{fig:disentange}
\end{figure}

\section{Factorized Dataset Distillation}
\label{sec:factorzied dd}
In representation learning, although the image data are in the extremely high-dimensional space, they may lie on a low-dimensional manifold and rely on a few features, and one can recover the source image from the low-dimensional features with specific decoders \citep{bengio2013representation,zhang2018network}. For this reason, it is plausible to implicitly learn synthetic datasets by optimizing their factorized features and corresponding decoders, which is termed as {\it factorized dataset distillation}, as shown in Figure \ref{fig:disentange}. To be in harmony with decoders, we recall the notion of a feature as code in terms of the dataset. By factorizing synthetic datasets with a combination of codes and decoders, the compression ratio can be further decreased, and information redundancy in distilled images can also be reduced. According to the learnability of the code and decoder, we classify factorized dataset distillation into three categories of code-based DD, decoder-based DD, and code-decoder DD.

Code-based DD aims to learn a series of low-dimensional codes for generating highly informative images through a specific generator. In \citep{zhao2022synthesizing}, the vectors that are put into the GAN generator are learned to produce informative images. Concretely, these investigators inverse real examples with a GAN generator and collect corresponding latent codes. Then, the latent vectors are further optimized with the distribution matching algorithm. By this, the optimized latent vectors can induce more informative synthetic examples with the pretrained GAN generator. In addition to using the GAN, \citet{kim2022dataset} employed a deterministic multi-formation function $\texttt{multi-form}(\cdot)$ as the decoder to create synthetic data from fewer condensed data, {\it i.e.}, the synthetic data are generated by $\mathcal{S} = \texttt{multi-form} (\mathcal{C})$. Then the condensed data $\mathcal{C}$ is optimized in an end-to-end fashion by the gradient matching approach. {Besides, the Partitioning and Expansion augmentation strategy adopted in \citep{zhao2023improved} can also be regarded as a non-parametric decoder to generate more synthetic data based on the distilled codes. Concretely, while the distilled codes possess the same dimension as real images, \citet{zhao2023improved} firstly partition each code into several patches and then expand the small patches into the standard dimension during both dataset distillation and network training periods.}  {Recently, \citet{cazenavette2023glad} argued that this low cross-architecture transferability comes from direct optimization on the pixel space, which makes the distilled data overfit to specific architectures used in the DD process. With this argument, they regularized the distillation by optimizing the intermediate latent codes that are fed into pretrained generative models for synthetic data generation, which largely increases the cross-architecture performance due to the more realistic synthetic data.}

Different from code-based DD, decoder-based DD solely learns decoders that are used to produce highly informative data. In \citep{such2020generative}, a generative teaching network (GTN) that employs a trainable network was proposed to generate synthetic images from random noise based on given labels, and the meta-gradients are backpropagated via BPTT to update the GTN rather than the synthetic data.

Naturally, code-decoder DD combines code-based and decoder-based DD and allows the training on both codes and decoders. \citet{deng2022remember} generated synthetic images via the matrix multiplication between codes and decodes, which they call the {\it memory} and {\it addressing function}. Specifically, they use the memory $\mathcal{M} = \{\boldsymbol{b}_1, \cdots, \boldsymbol{b}_K\}$ to store the bases $\boldsymbol{b}_i \in \mathbb{R}^{d}$ that have the same dimension as target data; and $r$ addressing functions of $\mathcal{A} = \{\boldsymbol{A}_1,\cdots,\boldsymbol{A}_r\}$ are used to recover the synthetic images according to the given one-hot label $\boldsymbol{y}$ as follows:
\begin{equation}
\label{eq:memory}
    \boldsymbol{s}_i^\top = \boldsymbol{y}^T \boldsymbol{A}_i [\boldsymbol{b}_1, \cdots, \boldsymbol{b}_K] ^\top,
\end{equation}
where the one-hot label $\boldsymbol{y} \in \mathbb{R}^{C\times 1}$, and $\boldsymbol{A}_i \in \mathbb{R}^{C\times K}$, and $C$ is the number of classes. By plugging different addressing functions $\boldsymbol{A}_i$ into Eq. \ref{eq:memory}, a total number of $r$ synthetic images can be generated from the bases for each category. During the distillation process, these two elements of $\mathcal{M}$ and $\mathcal{A}$ were optimized with the meta-learning framework. Different from matrix multiplication in \citep{deng2022remember}, \citet{liu2022dataset} employ {\it hallucinator} networks as decoders to generate synthetic data from {\it basis}. Specifically, a hallucinator network consists of three parts of an encoder $\texttt{enc}$, an affine transformation with trainable scale $\sigma$ and shift $\mu$, and a decoder $\texttt{dec}$, then a basis $\boldsymbol{b}$ is put into the hallucinator network to generate corresponding synthetic image $\boldsymbol{s}$ as follows:
\begin{equation}
    \mathbf{f}_1 = \texttt{enc}(\boldsymbol{b}), \quad \mathbf{f}_2 = \sigma \times \mathbf{f}_1 + \mu, \quad \boldsymbol{s} = \texttt{dec}(\mathbf{f}_2),
\end{equation}
    where the multiplication is element-wise. In addition, to enlarge the knowledge divergence encoded by different hallucinator networks for efficient compression, these researchers proposed an adversarial contrastive constraint that maximizes the divergence of features of different generated synthetic data. A similar code-decoder factorization method is also presented in \citet{lee2022kfs}, where they adopted an improved distribution matching to optimize the latent codes and decoders. {Recently, \citet{zhang2023dataset} employ the powerful generative model as the decoder and optimize it and corresponding latent codes. Specifically, each low-dimensional latent code can generate synthetic images with different classes by feeding the generative model with corresponding conditional class embedding. By this, the number of learnable parameters is not affected by both the real image resolution and the number of classes, which is promising for tackling the large-scale distillation problem.}

Through implicitly learning the latent codes and decoders, factorized DD possesses advantages such as more compact representation and shared representation across classes that consequently improve the dataset distillation performance. Notably, this code-decoder factorization is compatible with the aforementioned distillation approaches. Therefore, the exploration of synthetic data generation and distillation frameworks can promote dataset distillation in parallel.

\section{Performance Comparison}
\label{sec:performance comparison}

\begin{table*}[t]
    
    \centering
    \caption{Performance comparison of different dataset distillation methods on different datasets. Abbreviations of GM, TM, and DM are for gradient matching, trajectory matching, and distribution matching, respectively. Whole denotes the test accuracy in terms of the whole target dataset.  {The factorized DD methods are marked by $^\ast$.}}
    \resizebox{\linewidth}{!}{
    \begin{tabular}{c|c|c|c|c|c|c|c|c|c|c|c|c|c|c|c|c|c|c|c}
    \toprule

    \multirow{2}{*}{Methods} & \multirow{2}{*}{Schemes} & \multicolumn{3}{c|}{MNIST} & \multicolumn{3}{c|}{FashionMNIST} & \multicolumn{3}{c|}{SVHN} & \multicolumn{3}{c|}{CIFAR-10} & \multicolumn{3}{c|}{CIFAR-100} & \multicolumn{3}{c}{Tiny ImageNet} \\
    \cline{3-20}
     & & $1$ & $10$ & $50$ & $1$ & $10$ & $50$ & $1$ & $10$ & $50$ & $1$ & $10$ & $50$ & $1$ & $10$ & $50$ & $1$ & $10$ & $50$ \\

    \hline
    Random & - & \format{64.9}{3.5} & \format{95.1}{0.9} & \format{97.9}{0.2}  & \format{51.4}{3.8} & \format{73.8}{0.7} & \format{82.5}{0.7} & \format{14.6}{1.6} & \format{35.1}{4.1} & \format{70.9}{0.9} & \format{14.4}{2.0} & \format{26.0}{1.2} & \format{43.4}{1.0}  & \format{4.2}{0.3} & \format{14.6}{0.5} & \format{30.0}{0.4} & \format{1.4}{0.1} & \format{5.0}{0.2} & \format{15.0}{0.4} \\

    Herding & - & \format{89.2}{1.6} & \format{93.7}{0.3} & \format{94.8}{0.2}  & \format{67.0}{1.9} & \format{71.1}{0.7} & \format{71.9}{0.8} & \format{20.9}{1.3} & \format{50.5}{3.3} & \format{72.6}{0.8} & \format{21.5}{1.2} & \format{31.6}{0.7} & \format{40.4}{0.6}  & \format{8.4}{0.3} & \format{17.3}{0.3} & \format{33.7}{0.5} & \format{2.8}{0.2} & \format{6.3}{0.2} & \format{16.7}{0.3} \\

    DD \citep{wang2018dataset} & BPTT & - & \format{79.5}{8.1} & -  & - & - & - &- & - & - & 
    - & \format{36.8}{1.2} & -  & - & - & - & 
    - & - & - \\

    LD \citep{bohdal2020flexible} & BPTT & \format{60.9}{3.2} & \format{87.3}{0.7} & \format{93.3}{0.3}  & 
   - & - & - & - & - & - & 
   \format{25.7}{0.7} & \format{38.3}{0.4} & \format{42.5}{0.4}  & \format{11.5}{0.4} & - & - & 
   - & - & - \\

    DC \citep{zhao2021dataset} & GM & \format{91.7}{0.5} & \format{94.7}{0.2} & \format{98.8}{0.2}  & 
    \format{70.5}{0.6} & \format{82.3}{0.4} & \format{83.6}{0.4} & \format{31.2}{1.4} & \format{76.1}{0.6} & \format{82.3}{0.3} & \format{28.3}{0.5} & \format{44.9}{0.5} & \format{53.9}{0.5} & \format{12.8}{0.3} & \format{26.6}{0.3} & \format{32.1}{0.3} & - & - & - \\

    DSA \citep{zhao2021dsa} & GM & \format{88.7}{0.6} & \format{97.8}{0.1} & \orformat{99.2}{0.1}  & 
    \format{70.6}{0.6} & \format{86.6}{0.3} & \format{88.7}{0.2} & \format{27.5}{1.4} & \orformat{79.2}{0.5} & \format{84.4}{0.4} & \format{28.8}{0.7} & \format{52.1}{0.5} & \format{60.6}{0.5}  & \format{13.9}{0.4} & \format{32.4}{0.3} & \format{38.6}{0.3} & - & - & -\\

    DCC \citep{lee2022dataset} & GM & - & - & -  & 
    - & - & - & 
    \format{47.5}{2.6} & \orformat{80.5}{0.6} & \orformat{87.2}{0.3} & \format{34.0}{0.7} & \format{54.5}{0.5} & \format{64.2}{0.4}  & \format{14.6}{0.3} & \format{33.5}{0.3} & \format{39.3}{0.4} & - & - & - \\

    MTT \citep{cazenavette2022distillation} & TM & \format{91.4}{0.9} & \format{97.3}{0.1} & \format{98.5}{0.1}  & 
    \format{75.1}{0.9} & \orformat{87.2}{0.3} & \format{88.3}{0.1} & - & - & - & 
    \format{46.3}{0.8} & \format{65.3}{0.7} & \format{71.6}{0.2}  & \format{24.3}{0.3} & \format{40.1}{0.4} & \format{47.7}{0.2} & \format{8.8}{0.3} & \format{23.2}{0.2} & \orformat{28.0}{0.3} \\

    FTD \citep{du2022minimizing} & TM & - & - & -  & - & - & - & - & - & - & 
    \format{46.8}{0.3} & \orformat{66.6}{0.3} & \orformat{73.8}{0.2}  & \format{25.2}{0.2} & \orformat{43.4}{0.3} & \orformat{50.7}{0.3} & \format{10.4}{0.3} & \format{24.5}{0.2} & - \\

    TESLA \citep{cui2023scaling} & TM & - & - & -  & - & - & - & - & - & - & 
    \format{48.5}{0.8} & \format{66.4}{0.8} & \format{72.6}{0.7}  & \format{24.8}{0.4} & \format{41.7}{0.3} & \orformat{47.9}{0.3} & \format{7.7}{0.2} & \format{18.8}{1.3} & \orformat{27.9}{1.2} \\

    DM \citep{zhao2023distribution} & DM & \format{89.2}{1.6} & \format{97.3}{0.3} & \format{94.8}{0.2}  & 
   - & - & - & - & - & - & 
   \format{26.0}{0.8} & \format{48.9}{0.6} & \format{63.0}{0.4}  & \format{11.4}{0.3} & \format{29.7}{0.3} & \format{43.6}{0.4} & \format{3.9}{0.2} & \format{12.9}{0.4} & \format{24.1}{0.3} \\

    CAFE \citep{wang2022cafe} & DM & \format{90.8}{0.5} & \format{97.5}{0.1} & \format{98.9}{0.2}  & 
    \format{73.7}{0.7} & \format{83.0}{0.3} & \format{88.2}{0.3} & \format{42.9}{3.0} & \format{77.9}{0.6} & \format{82.3}{0.4} & \format{31.6}{0.8} & \format{50.9}{0.5} & \format{62.3}{0.4}  & \format{14.0}{0.3} & \format{31.5}{0.2} & \format{42.9}{0.2} & - & - & - \\

    KIP \citep{nguyen2020dataset,nguyen2021dataset} & KRR & \format{90.1}{0.1} & \format{97.5}{0.0} & \format{98.3}{0.1}  & 
    \format{73.5}{0.5} & \format{86.8}{0.1} & \format{88.0}{0.1} & \orformat{57.3}{0.1} & \format{75.0}{0.1} & \orformat{85.0}{0.1} & \format{49.9}{0.2} & \format{62.7}{0.3} & \format{68.6}{0.2}  & \format{15.7}{0.2} & \format{28.3}{0.1} & - & 
    - & - & - \\

    FRePo \citep{zhou2022dataset} & KRR & \format{93.0}{0.4} & \orformat{98.6}{0.1} & \orformat{99.2}{0.0}  & 
    \format{75.6}{0.3} & \format{86.2}{0.2} & \orformat{89.6}{0.1} & - & - & - & 
    \format{46.8}{0.7} & \format{65.5}{0.4} & \format{71.7}{0.2}  & \orformat{28.7}{0.1} & \format{42.5}{0.2} & \format{44.3}{0.2} & \orformat{15.4}{0.3} & \orformat{25.4}{0.2} & - \\

    RFAD \citep{loo2022efficient} & KRR & \orformat{94.4}{1.5} & \format{98.5}{0.1} & \format{98.8}{0.1}  & 
    \orformat{78.6}{1.3} & \format{87.0}{0.5} & \format{88.8}{0.4} & \orformat{52.2}{2.2} & \format{74.9}{0.4} & \format{80.9}{0.3} & \orformat{53.6}{1.2} & \format{66.3}{0.5} & \format{71.1}{0.4}  & \format{26.3}{1.1} & \format{33.0}{0.3} & - &
    - & - & - \\

    RCIG \citep{loo2023dataset} & KRR & \orformat{94.7}{0.5} & \orformat{98.9}{0.0} & \orformat{99.2}{0.0}  & 
    \orformat{79.8}{1.1} & \orformat{88.5}{0.2} & \orformat{90.2}{0.2} & - & - & - & \orformat{53.9}{1.0} & \orformat{69.1}{0.4} & \orformat{73.5}{0.3}  & \orformat{39.3}{0.4} & \orformat{44.1}{0.4} & \format{46.7}{0.3} & \orformat{25.6}{0.3}
     & \orformat{29.4}{0.2} & - \\

    IDC$^\ast$ \citep{kim2022dataset} & GM & \format{94.2}{0.2} & \format{98.4}{0.1} & \format{99.1}{0.1}  & \format{81.0}{0.2} & \format{86.0}{0.3} & \format{86.2}{0.2} & \format{68.1}{0.1} & \format{87.3}{0.2} & \format{90.2}{0.1} & \format{50.0}{0.4} & \format{67.5}{0.5} & \format{74.5}{0.1}  & - & \format{44.8}{0.2} & - &
    - & - & -\\

    DREAM$^\ast$ \citep{liu2023dream} & GM & \format{95.7}{0.3} & \format{98.6}{0.1} & \format{99.2}{0.1}  & \format{81.3}{0.2} & \format{86.4}{0.3} & \format{86.8}{0.3} & \format{69.8}{0.8} & \format{87.9}{0.4} & \format{90.5}{0.1} & \format{51.1}{0.3} & \format{69.4}{0.4} & \format{74.8}{0.1}  & \format{29.5}{0.3} & \format{46.8}{0.7} & \blformat{52.6}{0.4} &
    \format{10.0}{0.4} & \format{23.9}{0.4} & \blformat{29.5}{0.3}\\

    RTP$^\ast$ \citep{deng2022remember} & BPTT & \blformat{98.7}{0.7} & \blformat{99.3}{0.5} & \blformat{99.4}{0.4}  & 
    \blformat{88.5}{0.1} & \blformat{90.0}{0.7} & \blformat{91.2}{0.3} & \blformat{87.3}{0.1} & \format{89.1}{0.2} & \format{89.5}{0.2} & \blformat{66.4}{0.4} & \format{71.2}{0.4} & \format{73.6}{0.4}  & \format{34.0}{0.4} & \format{42.9}{0.7} & - & 
    \format{16.0}{0.7} & - & - \\

    HaBa$^\ast$ \citep{liu2022dataset} & TM & - & - & -  & - & - & - & \format{69.8}{1.3} & \format{83.2}{0.4} & \format{88.3}{0.1} & \format{48.3}{0.8} & \format{69.9}{0.4} & \format{74.0}{0.2}  & \format{33.4}{0.4} & \format{40.2}{0.2} & \format{47.0}{0.2} & - & - & - \\

    IDM$^\ast$ \citep{zhao2023improved} & DM & - & - & -  & - & - & - & 
    - & - & - & \format{45.6}{0.7} & \format{58.6}{0.1} & \format{67.5}{0.1}  & \format{20.1}{0.3} & \format{45.1}{0.1} & \format{50.0}{0.2} & 
    \format{10.1}{0.2} & \format{21.9}{0.2} & \format{27.7}{0.3} \\

    KFS$^\ast$ \citep{lee2022kfs} & DM & - & - & -  & - & - & - & 
    \format{82.9}{0.4} & \blformat{91.4}{0.2} & \blformat{92.2}{0.1} & \format{59.8}{0.5} & \blformat{72.0}{0.3} & \blformat{75.0}{0.2}  & \blformat{40.0}{0.5} & \blformat{50.6}{0.2} & - & 
    \blformat{22.7}{0.3} & \blformat{27.8}{0.2} & - \\
   \cline{3-20}
   Whole & - & \multicolumn{3}{c|}{\format{99.6}{0.0}} & \multicolumn{3}{c|}{\format{93.5}{0.1}} & \multicolumn{3}{c|}{\format{95.4}{0.1}} & \multicolumn{3}{c|}{\format{84.8}{0.1}} & \multicolumn{3}{c|}{\format{56.2}{0.3}} & \multicolumn{3}{c}{\format{37.6}{0.4}}\\
    \bottomrule
    \end{tabular}
    }
    \label{tab:dd performance}
\end{table*}

\begin{table}[t]
    \centering
    \caption{Performance comparison of different dataset distillation methods on ImageNet-1K.}
    \begin{tabular}{c|c|c|c | c }
    \toprule
    Methods & IPC=$1$ & IPC = $2$ & IPC = $10$ & IPC = $50$  \\
    \hline
    Random & \format{0.5}{0.1}  & \format{0.9}{0.1} & \format{3.6}{0.1} & \format{15.3}{2.3} \\
    DM \citep{zhao2023distribution} & \format{1.5}{0.1} & \format{1.7}{0.1} & - & - \\

    FRePo \citep{zhou2022dataset} & \format{7.5}{0.3} & \format{9.7}{0.2} & - & - \\
    TESLA \citep{cui2023scaling} & \format{7.7}{0.2} & \format{10.5}{0.3} & \bformat{17.8}{1.3} & \bformat{27.9}{1.2} \\
    RCIG \citep{loo2023dataset} & \bformat{15.6}{0.2} & \bformat{16.6}{0.1} & - & - \\
    \cline{2-5}
    Whole & \multicolumn{4}{c}{\format{33.8}{0.3}}  \\
    \bottomrule
    \end{tabular}
    \label{tab:dd performance imagenet}
\end{table}

To demonstrate the effectiveness of dataset distillation, we collect and summarize the classification performance of some characteristic dataset distillation approaches on the following image datasets: MNIST \citep{lecun1998gradient}, FashionMNIST \citep{xiao2017/online} , SVHN \citep{netzer2011reading}, CIFAR-10/100 \citep{krizhevsky2009learning}, Tiny ImageNet \citep{le2015tiny}, and ImageNet-1K \citep{russakovsky2015imagenet}. The details of these datasets are presented as follows. Recently, \citet{cui2022dc} publish a benchmark in terms of dataset distillation, however, it only contains five DD methods, and we provide a comprehensive comparison of over $15$ existing dataset distillation methods in this survey.

MNIST is a black-and-white dataset that consists of $60,000$ training images and $10,000$ test images from $10$ different classes; the size of  each example image is $28\times 28$ pixels. SVHN is a colorful dataset and consists of $73,257$ digits and $26,032$ digits for training and testing, respectively, and the example images in SVHN are $32\times 32$ RGB images. CIFAR-10/100 are composed of $50,000$ training images and $10,000$ test images from $10$ and $100$ different classes, respectively. The RGB images in CIFAR-10/100 are comprised of $32\times 32$ pixels. Tiny ImageNet consists of $100,000$ and $10,000$ $64\times 64$ RGB images from $200$ different classes for training and testing, respectively. ImageNet-1K is a large image dataset that consists of over $1,000,000$ high-resolution RGB images from $1,000$ different classes.

An important factor affecting the test accuracy {\it w.r.t.} the distilled dataset is the {\it distillation budget}, which constrains the size of the distilled dataset by the notion of images allocated per class (IPC). Usually, the distilled dataset is set to have the same number of classes as the target dataset. Therefore, for a target dataset with $100$ classes, setting the distillation budget to $\text{IPC}=10$ suggests that there are a total of $10\times 100 = 1,000$ images in the distilled dataset. 

\begin{table*}
    \centering
    \caption{Cross-architecture performance {\it w.r.t.} ResNet (RN), DenseNet (DN), and ViT on CIFAR-10. The factorized DD methods are marked by $^\ast$.} 
    \resizebox{\linewidth}{!}{
    \begin{tabular}{c|cccccc|cccccc|cccccc}
    \toprule
    \multirow{2}{*}{Methods} & \multicolumn{6}{c|}{IPC=$1$} & \multicolumn{6}{c|}{IPC=$10$} & \multicolumn{6}{c}{IPC=$50$}   \\ 
    \cline{2-19}
     & \multicolumn{1}{c}{ConvNet} & \multicolumn{1}{c}{RN-10} & \multicolumn{1}{c}{RN-18} & \multicolumn{1}{c}{RN-152} & \multicolumn{1}{c}{DN-121} & \multicolumn{1}{c|}{ViT} & \multicolumn{1}{c}{ConvNet} & \multicolumn{1}{c}{RN-10} & \multicolumn{1}{c}{RN-18} & \multicolumn{1}{c}{RN-152} & \multicolumn{1}{c}{DN-121} & \multicolumn{1}{c|}{ViT} & \multicolumn{1}{c}{ConvNet} & \multicolumn{1}{c}{RN-10} & \multicolumn{1}{c}{RN-18} & \multicolumn{1}{c}{RN-152} & \multicolumn{1}{c}{DN-121} & \multicolumn{1}{c}{ViT} \\
    \hline
    DC \citep{zhao2021dataset} & \format{28.3}{0.5} & - & \format{25.6}{0.6} & \orformat{15.3}{0.4} & - &  \orformat{24.68}{0.7}  &  \format{44.9}{0.5} & - & \format{42.1}{0.6} & \format{16.1}{1.0}& - & \format{28.96}{0.6} & \format{53.9}{0.5}&  - & \format{45.9}{1.4}& \format{19.7}{1.2} & - & \format{27.61}{0.3}\\
    
    DSA \citep{zhao2021dsa} & \format{28.8}{0.5} &  \format{25.1}{0.8} & \format{25.6}{0.6} &  \format{15.1}{0.7} & \format{25.9}{1.8} & \format{23.69}{0.8} & \format{52.1}{0.5} & \format{31.4}{0.9} & \format{42.1}{0.6}& \format{16.1}{1.0} & \format{32.9}{1.0} & \format{31.9}{0.4} &   \format{60.6}{0.5} &  \format{49.0}{0.7} & \format{49.5}{0.7} & \format{20.0}{1.2} & \format{53.4}{0.8} & \format{43.19}{0.7} \\

    MTT \citep{cazenavette2022distillation} & \format{46.3}{0.8}& - &  \orformat{34.2}{1.4} & \format{13.4}{0.9} & - & \format{21.67}{0.5} & \format{65.3}{0.7} & - & \format{38.8}{0.7} & \format{15.9}{0.2}&-  & \format{34.6}{0.6} &  \format{71.6}{0.2}& - & \format{60.0}{0.7}& \format{20.9}{1.6} & - & \orformat{48.00}{0.3}\\

    %FTD \citep{du2022minimizing}&- & - & - & -& - & - &-&-&-&  - & - & - & \format{73.8}{0.2}& - &\orformat{65.7}{0.3}  & - & -  & - \\

    TESLA \citep{cui2023scaling} & - &-& -& -&- & - & \format{66.4}{0.8}& - & \orformat{48.9}{2.2}& - & - & \orformat{34.8}{1.2}& -&-&- & - & - & -\\

    DM \citep{zhao2023distribution}& \format{26.0}{0.8}& \format{13.7}{1.6} & \format{20.6}{0.5} & \format{14.1}{0.6} & \format{12.9}{1.8} & \format{21.34}{0.2} &   \format{48.9}{0.6}& \format{31.7}{1.1} & \format{38.2}{1.1}& \format{15.6}{1.5}& \format{32.2}{0.8} & \format{34.4}{0.5} & \format{63.0}{0.4}& \format{49.1}{0.7} & \format{52.8}{0.4}& \orformat{21.7}{1.3} & \format{53.7}{0.7} & \format{45.07}{0.5}\\

    %CAFE \citep{wang2022cafe}&- & - & - & -& -&-&-& -&-&  - & - & - & \format{55.5}{0.4}& \format{25.3}{0.9}& - & - & - & - \\
    
    KIP \citep{nguyen2021dataset} & \format{49.9}{0.2} & - & \format{27.6}{1.1} & \format{14.2}{0.8} & - & \format{16.80}{0.8} &  \format{62.7}{0.3}& - & \format{45.2}{1.4} & \orformat{16.6}{1.4}& - & \format{15.9}{1.1} & \format{68.6}{0.2}& - & \format{60.0}{0.7}& \format{20.9}{1.6} & - & \format{18.56}{0.84} \\

    IDC$^\ast$ \citep{kim2022dataset}&\format{50.0}{0.4} & \format{41.9}{0.6} & - & -& \format{39.8}{1.2} & - & \format{67.5}{0.5}& \format{63.5}{0.1}&-&  - & \format{61.6}{0.6} & - & \format{74.5}{0.1}& \format{72.4}{0.5} & - & - & \format{71.8}{0.6} & - \\

    KFS$^\ast$ \citep{lee2022kfs}& \orformat{59.8}{0.5} & \orformat{47.0}{0.8} & - & -& \orformat{49.5}{1.3}  & - & \orformat{72.0}{0.3}& \orformat{70.3}{0.3} & -&  - & \orformat{71.4}{0.4} & - & \orformat{75.0}{0.2}& \orformat{75.1}{0.3}& - & - & \orformat{76.3}{0.4} & - \\
    \bottomrule
    \end{tabular}
    }
    \label{tab:cross architecture 1}
\end{table*}

\subsection{Standard Benchmark}
\label{sec:standard benchmark}

For a comprehensive comparison, we also present the test accuracy {\it w.r.t.} a random select search, coreset approach, and the original target dataset. For the coreset approach, the Herding algorithm \citep{rebuffi2017icarl} is employed in this survey due to its superior performance. Notably, for most parametric DD methods, ConvNet \citep{zhao2021dataset, gidaris2018dynamic}, {which consists of three convolutional blocks}, is the default architecture to distill the synthetic data; and all the test accuracy of distilled data are obtained by training the same architecture of ConvNet for a fair comparison. 

The empirical comparison results associated with MNIST, FashionMNIST, SVHN, CIFAR-10/100, and Tiny ImageNet are presented in Table \ref{tab:dd performance}. In spite of severe scale issues and few implementations on ImageNet-1K, we present the DD results of ImageNet-1K in Table \ref{tab:dd performance imagenet} for completion. As shown in these tables, many methods are not tested on some datasets due to the lack of meaning for easy datasets or scalability problems for large datasets. We preserve these blanks in tables for a more fair and comprehensive comparison. To make the comparison clear, we use orange numbers for the best of two nonfactorized DD methods and blue numbers for the best factorized DD methods in each category in Table \ref{tab:dd performance}. In addition, we note the factorized DD methods with $^\ast$. Based on the performance comparison in the tables, we observe the following:
\begin{itemize}
    \item Dataset distillation can be realized on many datasets with various sizes.
    \item Dataset distillation methods outperform random pick and coreset selection by large margins.
    \item KRR and trajectory matching approaches possess advanced DD performance {\it w.r.t.} three larger datasets of CIFAR-10/100 and Tiny ImageNet among nonfactorized DD methods.
    \item Factorized dataset distillation by optimizing the latent codes and decoders can largely improve the test accuracy of distilled data.
    \item KFS \citep{lee2022kfs} has the best overall performance among different datasets of CIFAR-10/100 and Tiny ImageNet.
\end{itemize}

\subsection{Cross-Architecture Transferability}
\label{sec:cross architecture transferability}

The standard benchmark in Table \ref{tab:dd performance} only shows the performance of DD methods in terms of the specific ConvNet, while the distilled dataset is usually used to train other unseen network architectures. Therefore, measuring the DD performance on different architectures is also important for a more comprehensive evaluation. However, there is not a standard set of architectures to evaluate the cross-architecture transferability of DD algorithms, and the choice of architectures is highly different in the DD literature. 

In this survey, we collect {synthetic data distilled by different methods on ConvNet to train other} five popular architectures of ResNet-10/18/152 \citep{he2016deep}, DenseNet-121 \citep{huang2017densely}, {and Vision Transformer (ViT) \footnote{{The ViT architecture has $4\times 4$ patch resolution, $6$ layers, $8$ heads, hidden dimension of $512$, and the MLP dimension of $512$.} } \citep{dosovitskiy2021an}} to evaluate the cross-architecture transferability of distilled data, as shown in Tables \ref{tab:cross architecture 1}. %and \ref{tab:cross architecture 2}. 
By investigating these tables, we can obtain several observations as follows:

\begin{itemize}
    \item There is a significant performance drop for nonfactorized DD methods when the distilled data are used to train unseen architectures.
    \item The distilled data do not always possess the best performance across different architectures, which underlines the importance of evaluation on multiple architectures.
    \item The factorized DD methods (IDC and KFS) possess better cross-architecture transferability, {\it i.e.}, suffer a smaller accuracy drop when encountering unseen architectures.
\end{itemize}

\section{Data Modalities}
Apart from image data that the majority of DD works focus on, some works leverage dataset distillation on multimodal data such as graph, text, {\it etc.}

\subsection{Graph Data}
Regarding a graph dataset of $\mathcal{T}=\{\mathbf{A}, \mathbf{X}, \mathbf{Y}\}$ with $N$ nodes, where $\mathbf{A}$, $\mathbf{X}$, and $\mathbf{Y}$ are the adjacency matrix, the node feature matrix, and the node labels, respectively, a graph neural network $\mathcal{G}$ is trained on $\mathcal{T}$ for node classification. \citet{jin2022graph} adopted the gradient matching approach to distill the graph $\mathcal{T}$ to a small synthetic graph $\mathcal{S}=\{\mathbf{A}^\prime, \mathbf{X}^\prime, \mathbf{Y}^\prime\}$ with $N^\prime$ nodes and $N^\prime \ll N$.

However, the number of parameters in the adjacency matrix $\mathbf{A}^\prime$ grows quadratically as $N^\prime$ increases, which greatly impedes the scalability of graph distillation. Considering the implicit correlation between the graph structure and node features, the authors employ an MLP-based model to predict the graph structure based on node features:
\begin{equation}
    \mathbf{A}_{i j}^{\prime}=\sigma\left(\frac{\operatorname{MLP}_{\Phi}\left(\left[\mathbf{x}_i^{\prime} ; \mathbf{x}_j^{\prime}\right]\right)+\operatorname{MLP}_{\Phi}\left(\left[\mathbf{x}_j^{\prime} ; \mathbf{x}_i^{\prime}\right]\right)}{2}\right),
\end{equation}
where $\sigma(\cdot)$ is the sigmoid function, $\text{MLP}_{\Phi}$ is parameterized with $\Phi$, and $[\cdot;\cdot]$ denotes concatenation. With fixing $\mathbf{Y}^\prime$ to ease the difficulty of optimization, the loss function $\mathcal{L}(\mathbf{X}^\prime, \Phi)$ of graph distillation for node classification is defined as
\begin{equation}
    \sum_{t=0}^{T-1} D\left(\nabla_{\boldsymbol{\theta}} \mathcal{R}\left(\mathcal{G}_{\boldsymbol{\theta}_t}\left({\Phi}\left(\mathbf{X}^{\prime}\right), \mathbf{X}^{\prime}\right), \mathbf{Y}^{\prime}\right), \nabla_{\boldsymbol{\theta}} \mathcal{R}\left(\mathcal{G}_{\boldsymbol{\theta}_t}(\mathbf{A}, \mathbf{X}), \mathbf{Y}\right)\right).
\end{equation}
Moreover, to bypass second-order meta-gradient computation, \citet{liu2022dataset} accelerated the graph distillation through distribution matching from the perspective of receptive fields, which present the local graph {\it w.r.t.} a specific node.

However, for the graph classification problem where the adjacency matrix is discrete, the conventional DD algorithms are no longer applicable. To distill synthetic graphs with discrete adjacency matrix $\mathbf{A}^\prime\in \{0,1\}^{|\mathbf{A}^\prime|}$, \citet{jin2022condensing} formulate $\mathbf{A}^\prime$ as a probabilistic graph with Bernoulli distribution
\begin{equation}
    \mathbf{P}_{\Omega_{i j}}\left(\mathbf{A}_{i j}^{\prime}\right)=\mathbf{A}_{i j}^{\prime} \sigma\left(\Omega_{i j}\right)+\left(1-\mathbf{A}_{i j}^{\prime}\right) \sigma\left(-\Omega_{i j}\right),
\end{equation}
where $\sigma$ is the sigmoid function and $\Omega_{ij}\in \mathbb{R}$ is the learnable parameter. Then with the reparameterization trick, the parameter $\Omega$ becomes differentiable and $\mathbf{A}^\prime$ is accordingly synthesized by optimizing $\Omega$.

\subsection{Text Data}

For the discrete text data, \citet{li2021text} converted each piece of text into a continuous embedding matrix to adjust DD algorithms, and then a few high-informative synthetic embedding matrices were distilled for text classification. To fine-tune large language models like BERT \citep{devlin2018bert} with the distilled text data, \citet{maekawa2023text} further optimized the soft label by decreasing the KL divergence between the attention probabilities of distilled data and the model during the DD process, which facilitates efficient knowledge transfer to transformer models.

\section{Application}
\label{sec:application}
Due to this superior performance in compressing massive datasets, dataset distillation has been widely employed in many application domains that limit training efficiency and storage, including continual learning and neural architecture search. Furthermore, due to the correspondence between examples and gradients, dataset distillation can also benefit privacy preservation, federated learning, and adversarial robustness. In this section, we briefly review these applications {\it w.r.t} dataset distillation.

\subsection{Continual Learning}

During to the training process, when there is a shift in the training data distribution, the model will suddenly lose its ability to predict the previous data distribution. This phenomenon is referred to as {\it catastrophic forgetting} and is common in deep learning. To overcome this problem, continual learning has been developed to incrementally learn new tasks while preserving performance on old tasks \citep{rebuffi2017icarl,castro2018end}. A common method used in continual learning is the replay-based strategy, which allows a limited memory to store a few training examples for rehearsal in the following training. Therefore, the key to the replay-based strategy is to select highly informative training examples to store. Benefiting from extracting the essence of datasets, the dataset distillation technique has been employed to compress data for memory with limited storage \citep{zhao2021dataset,carta2022distilled}. Because the incoming data have a changing distribution, the frequency is high for updating the elements in memory, which leads to strict requirements for the efficiency of dataset distillation algorithms. To conveniently embed dataset distillation into the replay-based strategy, \citet{wiewel2021condensed} and \citet{sangermano2022sample} decomposed the process of synthetic data generation by linear or nonlinear combination, and thus fewer parameters were optimized during dataset distillation. \citet{gu2023ssd} develop the dynamic memory to maintain both distilled and real data, which helps better summarize data information with the distilled data. 

In addition to enhancing replay-based methods, dataset distillation can also learn a sequence of stable datasets, and the network trained on these stable datasets will not suffer from catastrophic forgetting \citep{masarczyk2020reducing}. % detail ...

\subsection{Neural Architecture Search}

For a given dataset, the technique of neural architecture search (NAS) aims to find an optimal architecture from thousands of network candidates for better generalization. The NAS process usually includes training the network candidates on a small proxy of the original dataset to save training time, and the generalization ranking can be estimated according to these trained network candidates. Therefore, it is important to design the proxy dataset such that models trained on it can reflect the model's true performance in terms of the original data, but the size of the proxy dataset requires control for the sake of efficiency. To construct proxy datasets, conventional methods have been developed, including random selection or greedy search, without altering the original data \citep{c2022speeding,li2020sgas}. \citet{such2020generative} first proposed optimizing a highly informative dataset as the proxy for network candidate selection. More subsequent works have considered NAS as an auxiliary task for testing the proposed dataset distillation algorithms \citep{zhao2021dataset,zhao2021dsa,zhao2023distribution,du2022minimizing,cui2022dc}. Through simulation on the synthetic dataset, a fairly accurate generalization ranking can be collected for selecting the optimal architecture while considerably reducing the training time.

\subsection{Federated Learning}

Federated learning has received increasing attention during training neural networks in the past few years due to its advantages in distributed training and private data protection \citep{yang2019federated}. The federated learning framework consists of multiple clients and one central server, and each client possesses exclusive data for training the corresponding local model \citep{mcmahan2017communication}. In one round of federated learning, clients transmit the induced gradients or model parameters to the server after training with their exclusive data. Then the central server aggregates the received gradients or parameters to update the model parameters and broadcast new parameters to clients for the next round. In the federated learning scenario, the data distributed in different clients are often non-i.i.d data, which causes a biased minimum {\it w.r.t.} the local model and significantly hinders the convergence speed. Consequently, the data heterogeneity remarkably imposes a communication burden between the server and clients. \citet{goetz2020federated,wang2023fed} proposed distilling a small set of synthetic data from original exclusive data by matching gradients, and then the synthetic data instead of a large number of gradients were transmitted to the server for model updating. Other distillation approaches such as BPTT  \citep{hu2022fedsynth,zhou2020distilled}, KRR \citep{song2022federated}, and distribution matching \citep{xiong2023feddm}, were also employed to compress the exclusive data to alleviate the communication cost at each transition round. However, \citet{liu2022meta} discovered that synthetic data generated by dataset distillation are still heterogeneous. To address the problem, these researchers proposed two strategies of dynamic weight assignment and meta knowledge sharing during the distillation process, which significantly accelerate the convergence speed of federated learning. Apart from compressing the local data, \citet{pi2022dynafed} also distilled data via trajectory matching in the server, which allows the synthetic data to possess global information. Then the distilled data can be used to fine-tune the server model for convergence speed up. 

\subsection{Other Applications}

Apart from the aforementioned applications, we also summarize other applications in terms of dataset distillation as follows.

Because of their small size, synthetic datasets have been applied to explainability algorithms \citep{loo2022efficient}. In particular, due to the small size of synthetic data, it is easy to measure how the synthetic examples influence the prediction of test examples. If the test and training images both rely on the same synthetic images, then the training image will greatly influence the prediction of the test image. In other words, the synthetic set becomes a bridge to connect the training and testing examples. 

Leveraging the characteristic of capturing the essence of datasets, \citet{cazenavette2022textures,chen2022fashion} adopted dataset distillation for visual design. Specifically, \citet{cazenavette2022textures} generated the representative textures by randomly cropping the synthetic images during the distillation process. In addition to extracting texture,  \citet{chen2022fashion} imposed the synthetic images to model the outfit compatibility through dataset distillation.

Due to their small size and abstract visual information, distilled data can also be applied in medicine, especially in medical image sharing \citep{li2023sharing}. Empirical studies on gastric X-ray images have shown the advantages of DD in medical image transition and the anonymization of patient information \citep{li2020soft,li2022compressed}. Through dataset distillation, hospitals can share their valuable medical data at lower cost and risk to collaboratively build powerful computer-aided diagnosis systems.

\section{Challenges and Future Directions}
\label{sec:challenges}
Due to its superiority in compressing datasets and training speedup, dataset distillation has promising prospects in a wide range of areas. In the following, we discuss existing challenges in terms of dataset distillation. Furthermore, we investigate plausible directions and provide insights into dataset distillation to promote future studies.

\subsection{Challenges}
 
\textbf{Scalability.} Although many approaches have been proposed to decrease memory utilization \citep{cui2023scaling} and accelerate convergence \citep{liu2023dream} for efficient DD, the scalability of DD is still a serious issue due to the hardness of optimizing numerous hyperparameters ({\it e.g.}, pixels for image data). For example, when distilling the $1000$-class ImageNet with the input resolution of $224\times 224$ and IPC of $50$, totally $224\times 224 \times 1000 \times 50 \approx 2.5 \text{billon}$ hyperparameters (pixels) are required to simultaneously optimize. Therefore, how to effectively search a feasible solution over the extremely high-dimensional hyperparameter space is important to scale DD algorithms to more complicated data ({\it e.g.}, image data with higher resolution, 3D data).

\textbf{Multimodal dataset distillation.} Whereas DD algorithms have achieved remarkable progress in terms of image \citep{wang2018dataset,zhao2021dataset}, graph \citep{jin2022graph}, text \citep{li2021text}, tabular \citep{medvedev2020tabular}, and recommendation system data \citep{sachdeva2022data}, their applications on other data modalities are still underexplored. Especially for the modalities such as video that rich spatial and temporal information are tangled with each other in a single data \citep{herath2017going}, it is still challenging for DD algorithms to simultaneously distill the spatial and temporal features into a small volume of data.  

\textbf{DD with complex labels.} The majority of DD techniques lie in the single-label classification task where its label space is an integer set. As for other challenging tasks such as detection \citep{ren2015faster} or segmentation \citep{minaee2021image}, their label spaces are located in much higher dimensional spaces. Unlike DD with single-label classification that the specific number of synthetic images (IPC) are pre-assigned for each class, enumerated assignment is impractical when the label space is highly dimensional. Therefore, to represent the rich label space, high-dimensional labels of synthetic data require cautious adjusting, which significantly hinders the DD efficiency especially when jointly optimized with the input data.

\textbf{Theory of dataset distillation.} Although various DD algorithms have been emerging, there have been few investigations on discussing the theory behind dataset distillation. Nevertheless, developing the theory is extremely necessary and will considerably promote the dataset distillation by directly improving the performance and also suggesting the correct direction of development. For example, the theory can help derive a better definition of the dataset knowledge and consequently increase the performance. In addition, the theoretical correlation between the distillation budget (the size of synthetic datasets) and performance can provide an upper bound on the test error, which can offer a holistic understanding and prevent researchers from blindly improving the DD performance. Hence, a solid theory is indispensable to take the development of DD to the next stage.

\subsection{Future Directions}
Despite aforementioned challenges existing in DD, there are still many promising directions that are worth exploring to  {(1) enhance dataset distillation performance and (2) develop trustworthy dataset distillation.}

\textbf{Optimization objective of DD.} As discussed in Sections \ref{sec:factorzied dd} and \ref{sec:standard benchmark}, optimizing factorized codes and decoders that cooperatively generate synthetic data outperform the direct optimization by a large margin.   {This suggests that the selection of optimization objectives ({\it e.g.}, RGB pixels, latent codes and decoders) has a significant influence on the final distillation performance. Apart from optimization on original data or factorized spaces, \citet{wang2023dim} turned to optimize the generative model to synthesize unlimited informative data, and \citet{cazenavette2023glad} optimized the intermediate features of generative models to generate more realistic synthetic data for better cross-architecture generalization. Therefore, investigating optimization on proper information carriers, such as optimizing a few 3D data to summarize information of 2D images, is a promising direction for further DD performance improvement.}

\textbf{Model augmentation.} During the process of DD, neural networks within different training stages are involved to compute the distillation loss, and then the synthetic data are accordingly updated. This triggers a problem that distilled data are sensitive to network parameters, and causes unsatisfied DD performance when training new networks whose parameters have a large discrepancy to models used for calculating distillation loss. Therefore, the sensitivity to model parameters is closely related to DD performance. To alleviate this sensitivity, early works use a simple model augmentation strategy that conducts DD on networks with different initialization and training stages \citep{wang2018dataset,zhao2021dataset}. Recently, \citet{zhang2022accelerating} showed that employing early-stage models trained with a few epochs as the initialization can achieve better distillation performance, and they further proposed weight perturbation methods to efficiently generate early-stage models for calculating distillation loss. Therefore, it is important for dataset distillation to investigate model augmentation algorithms to reduce the sensitivity of distilled data to model parameters and further improve DD performance.

\textbf{Cross-architecture transferability.} Because the distillation of synthetic data is closely related to the specific network architecture (or kernels in the KRR approach), there is naturally a performance drop when distilled datasets are applied to train networks with other unseen architectures. Recently, \citet{cazenavette2023glad} argued that this low cross-architecture transferability comes from direct optimization on the pixel space, which makes the distilled data overfit to specific architectures used in the DD process. With this argument, they regularized the distillation by optimizing the latent codes that are fed into pretrained generative models for distilled data generation, which largely increases the cross-architecture performance due to more realistic synthetic data. A decent cross-architecture generalization is significant and allows synthetic data to train more complex models that are prohibitively used in DD process due to unbearable computing overhead.

\textbf{Private dataset distillation.} As the real data are involved in dataset distillation, it is necessary to analyze the distilled data from the perspective of privacy protection.

Theoretically, \citet{dong2022privacy} and \citet{zheng2023differentially} built the connection between dataset distillation and differential privacy (DP), and proved the superiority of dataset distillation in privacy preservation over conventional private data generation methods. However, \citet{carlini2022no} argued that there exist flaws in both experimental and theoretical evidence in \citep{dong2022privacy}: (1) membership inference baseline should include all training examples, while only $1\%$ of training points are used to match the synthetic sample size, which thus contributes to the high attack accuracy of baseline; and (2) the privacy analysis is based on an assumption stating that the output of learning algorithms follows a exponential distribution in terms of loss, while the assumption has already implied differential privacy. Empirically, \citet{chen2022privacy} employed gradient matching to distill private datasets by adding DP noise to the gradients of synthetic data,  {which was followed by \citet{vinaroz2023dpkip} who privatized KIP with DP-SGD to achieve better privacy-accuracy trade-off compared to \citep{chen2022privacy}. Therefore, a powerful private dataset distillation algorithm helps reassuringly publish distilled data and alleviate data abuse.}

\textbf{Robust dataset distillation.} It is significant to develop robust DD algorithms such that models trained on distilled data own satisfied robustness {\it w.r.t.} malicious attacks and ambiguous test data, or on the contrary, attack downstream models via manipulating the synthetic data.

To enhance adversarial robustness, \citet{tsilivis2022can} and \citet{wu2022towards} employed dataset distillation to extract the information of adversarial examples and generate robust datasets. Then standard network training on the distilled robust dataset is sufficient to achieve satisfactory robustness to adversarial perturbations, which substantially saves computing resources compared to expensive adversarial training. For backdoor attack, \citet{liu2023backdoor} considered injecting backdoor triggers into the small distilled dataset. Because the synthetic data possess a small size, direct trigger insertion is less effective and also perceptible. Therefore, these coworkers turned to insert triggers into the target dataset during the dataset distillation procedure. In addition, they propose to iteratively optimize the triggers during the distillation process to preserve the triggers' information for better backdoor attack performance.  {These defense and attack algorithms in terms of DD invoke the distillation of robust synthetic datasets, and models trained on them can better defend against various attacks.}

Recently, \citet{zhu2023rethinking} revealed that models trained on distilled data tend to output over-confident predictions and have a bad calibration, due to discarding semantically meaningful information. To tackle this problem, they propose to randomly mask the synthetic data to compel them to contain more semantically complete information during the distillation process. Apart from estimating proper uncertainty for in-distribution data, it is also important to detect out-of-distribution (OoD) examples for models deployed in the open world. \citet{ma2023towards} additionally synthesized outliers based on corruption transformations of real data, which are called pseudo-outliers. Then the synthetic outliers are encompassed to train downstream models for enhancing the capability of OoD detection. Therefore, it is significant for trustworthy learning to construct distilled datasets that can induce proper uncertainty {\it w.r.t.} both in-distribution and out-of-distribution test data.

\section*{Acknowledge}

We sincerely thank \href{https://github.com/Guang000/Awesome-Dataset-Distillation}{Awesome-Dataset-Distillation} for its comprehensive and timely DD publication summary. We also thank Can Chen and Guang Li for their kind feedback and suggestions.

\bibliographystyle{IEEEtranN}
\bibliography{ref}

\end{document}